%% file: main.tex
\documentclass[11pt]{article}

\usepackage[final]{acl}

\usepackage{times}
\usepackage{latexsym}

\usepackage[T1]{fontenc}

\usepackage[utf8]{inputenc}
\usepackage{microtype}
\usepackage{inconsolata}
\usepackage{graphicx}

\usepackage{hyperref}
\usepackage{booktabs}
\usepackage{tabularx}
\usepackage{tcolorbox}
\usepackage{listings}
\usepackage{enumitem}
\usepackage{multirow}
\usepackage{pifont}
\usepackage{nicematrix}
\usepackage{amsmath}
\usepackage{xspace}
\usepackage{float}
\usepackage[dvipsnames,table,xcdraw]{xcolor}
\usepackage{xtab}
 
\newcommand{\cmark}{\ding{51}}
\newcommand{\xmark}{\ding{55}}

\definecolor{actionblue}{HTML}{0066CC}

\newcommand{\task}[0]{CLID\xspace}
\newcommand{\dataset}[0]{\textsc{WikiCollide}\xspace}
\newcommand{\system}[0]{\textsc{CLAIRE}\xspace}
\newcommand{\feverous}[0]{\textsc{FEVEROUS}\xspace}

\newcommand{\supports}{%
\tcbox[colback=green!10,colframe=green!50!black,boxrule=0.3pt,arc=2pt,
top=-3.5pt,bottom=-3.5pt,left=0.5pt,right=0.5pt,
nobeforeafter,tcbox raise base]{\texttt{\strut Supports}}%
}

\newcommand{\refutes}{%
\tcbox[colback=red!10,colframe=red!50!black,boxrule=0.3pt,arc=2pt,
top=-3.5pt,bottom=-3.5pt,left=0.5pt,right=0.5pt,
nobeforeafter,tcbox raise base]{\texttt{\strut Refutes}}%
}

\newcommand{\nei}{%
\tcbox[colback=gray!10,colframe=gray!50!black,boxrule=0.3pt,arc=2pt,
top=-3.5pt,bottom=-3.5pt,left=0.5pt,right=0.5pt,
nobeforeafter,tcbox raise base]{\texttt{\strut Not Enough Information}}%
}

\newcommand{\action}[1]{%
\tcbox[colback=actionblue!20,colframe=actionblue!50!black,boxrule=0.3pt,arc=2pt,
top=-3.5pt,bottom=-3.5pt,left=0.5pt,right=0.5pt,
nobeforeafter,tcbox raise base]{\texttt{\strut #1}}%
}

\lstdefinestyle{prompt}{
    backgroundcolor=\color{white},
    basicstyle=\ttfamily\scriptsize,
    breaklines=true,
    commentstyle=\color{black}\bfseries,
    keywordstyle=\color{black},
    stringstyle=\color{black}\bfseries,
    showstringspaces=false,
    frame=single,
    captionpos=b,
    morestring=[s]{<}{>},
    morecomment=[l]{\#},
    morecomment=[l]{\#\#},
    morecomment=[l]{\#\#\#},
    moredelim = [s][\color{black}]{**}{**},
}

\definecolor{mediumgreen}{RGB}{60, 179, 113}
\lstdefinelanguage{Jinja2}{
  morekeywords={},
  sensitive=false,
  moredelim=[s][\color{black}]{\{}{\}},
  moredelim=[s][\color{black}]{\%}{\%},
  moredelim=[s][\color{mediumgreen}]{\{####}{####\}},
}

\title{
  Detecting Corpus-Level Knowledge Inconsistencies \\
  in Wikipedia with Large Language Models
}

\author{Sina J. Semnani \quad
  Jirayu Burapacheep \quad
  Arpandeep Khatua \quad \\
  \textbf{Thanawan Atchariyachanvanit} \quad
  \textbf{Zheng Wang} \quad
  \textbf{Monica S. Lam} \\
  Computer Science Department \\
  Stanford University, Stanford, CA \\
  \texttt{\{sinaj,jirayu,akhatua,thanawan,peterwz,lam\}@cs.stanford.edu} \\
}

\begin{document}
\maketitle

\input{0_abstract}
\input{1_intro}
\input{2_related_work}

\input{3_task}
\input{4_system}
\input{5_dataset}

\input{6_experiment}

\input{7_conclusion}
\input{8_other}

\bibliography{anthology, custom}

\appendix

\input{9_appendix}

\end{document}

%% file: 0_abstract.tex
\begin{abstract}
Wikipedia is the largest open knowledge corpus, widely used worldwide and serving as a key resource for training large language models (LLMs) and retrieval-augmented generation (RAG) systems. Ensuring its accuracy is therefore critical. But how accurate is Wikipedia, and how can we improve it?

We focus on \textit{inconsistencies}, a specific type of factual inaccuracy, and introduce the task of \emph{corpus-level inconsistency detection}. We present \system, an agentic system that combines LLM reasoning with retrieval to surface potentially inconsistent claims along with contextual evidence for human review. In a user study with experienced Wikipedia editors, 87.5\% reported higher confidence when using \system, and participants identified 64.7\% more inconsistencies in the same amount of time.

Combining \system with human annotation, we contribute \dataset, the first benchmark of real Wikipedia inconsistencies. Using random sampling with \system-assisted analysis, we find that at least 3.3\% of English Wikipedia facts contradict another fact, with inconsistencies propagating into 7.3\% of \feverous and 4.0\% of AmbigQA examples. Benchmarking strong baselines on this dataset reveals substantial headroom: the best fully automated system achieves an AUROC of only 75.1\%.

Our results show that contradictions are a measurable component of Wikipedia and that LLM-based systems like \system can provide a practical tool to help editors improve knowledge consistency at scale.~\footnote{Dataset and code are available at \url{https://github.com/stanford-oval/inconsistency-detection}.}
\end{abstract}

%% file: 1_intro.tex
\section{Introduction}
\label{sec:introduction}

Wikipedia is a widely used source of knowledge, attracting billions of monthly visitors~\citep{statista_wikipedia_traffic_2024}. Although initially criticized for reliability, the English Wikipedia later gained broad acceptance as a reputable source~\citep{economist_wikipedia_2021}. Beyond public use, it plays a central role in natural language processing (NLP) research: Wikipedia is used to train large language models (LLMs), provide ground truth for retrieval-augmented generation (RAG) systems~\citep{semnani-etal-2023-wikichat, lewis2020retrieval, guu2020realm, zhang-etal-2024-spaghetti}, and supply gold answers for question answering and fact verification.

Given this reliance, ensuring Wikipedia's accuracy is critical. We focus specifically on internal inconsistencies: contradictory facts within Wikipedia that indicate errors requiring correction through consultation of original sources. In a crowdsourced repository, inconsistencies can arise from outdated information, limited awareness of related content during editing, or simple human error.

The corpus's vast scale makes comprehensive verification challenging for both humans and automated tools.
While Wikipedia is often used to detect hallucinations in LLMs, we instead leverage LLMs to detect inconsistencies in a human-curated corpus.
Our contributions are as follows:

\textbf{We formalize the task of Corpus-Level Inconsistency Detection (\task).} Given a fact from a corpus, the goal is to identify at least one other fact within the same corpus that contradicts it. While inconsistency detection has been studied at the sentence-pair and document levels, corpus-level detection remains largely unexplored. Recent work examines inconsistencies between retrieved information and the internal knowledge of LLMs~\citep{su2024textttconflictbank, jin2024tugofwar, xie2023adaptive, wang2025retrievalaugmented}, but often relies on synthetic edits or focuses solely on temporal drift~\citep{marjanovic-etal-2024-dynamicqa}. Our task differs from traditional knowledge-intensive settings~\citep{petroni-etal-2021-kilt}, such as question answering~\citep{kwiatkowski-etal-2019-natural, joshi-etal-2017-triviaqa} and fact verification~\citep{thorne-etal-2018-fever, jiang-etal-2020-hover}, which typically assume corpus consistency: finding a single supporting or refuting evidence item is sufficient. This assumption breaks down when the corpus itself contains contradictions. Figure~\ref{fig:task} illustrates this distinction using an example from the \feverous dataset~\cite{aly-etal-2021-fact} and contrasts it with how \system addresses the same case.

\textbf{We propose \system (Corpus-Level Assistant for Inconsistency REcognition), a system for surfacing inconsistencies in large corpora.} To support non-expert users, \system finds and displays not only candidate contradictions but also disambiguating context and explanations of specialized terminology. It features an interactive interface implemented as a browser extension that surfaces potential inconsistencies to Wikipedia visitors. In a user study with eight experienced editors, participants identified 64.7\% more inconsistencies within the same amount of time when using \system than when using search engines.

\textbf{We provide the first lower bound on the inconsistency rate in the English Wikipedia and in two widely used Wikipedia-based NLP benchmarks.} Through manual verification of \system outputs, we estimate that approximately 3.3\% of facts in the English Wikipedia contradict other statements in the corpus. To our knowledge, this is the first systematic attempt to quantify corpus-level inconsistencies in Wikipedia. In AmbigQA~\cite{min-etal-2020-ambigqa}, 4.0\% of questions have answers that contradict other content in the same Wikipedia dump, challenging the dataset's assumption of unambiguous, unique answers. In \feverous, 7.3\% of claims labeled as \supports~are contradicted by other evidence within Wikipedia, undermining the standard assumption of corpus consistency in fact verification.

\textbf{We introduce \dataset, a dataset for Corpus-Level Inconsistency Detection on Wikipedia.} \dataset contains inconsistencies identified in the English Wikipedia. Unlike synthetic datasets, it captures genuine ambiguities and complex factual relationships arising in real-world content. To ensure meaningful coverage, we target Wikipedia's Level 5 Vital Articles,\footnote{\url{https://en.wikipedia.org/wiki/Wikipedia:Vital_articles/Level/5}} which are prioritized for improvement by WikiProject Vital Articles and serve as a centralized watchlist of important entries. Is finding inconsistencies in these pages like finding needles in a haystack? With \system-assisted curation, \dataset comprises 955 facts, 34.7\% of which are inconsistent.

\textbf{We evaluate \system and establish strong baselines on \dataset.} On the \dataset test set, \system achieves an AUROC of 75.1\%, outperforming baselines while leaving substantial headroom for future work.

%% file: 2_related_work.tex
\section{Related Work}
\label{sec:related_work}

\paragraph{Fact Verification.}
Recent advances in fact verification increasingly leverage LLMs~\citep{luu2024zefav, jayaweera-etal-2024-amrex}, often within retrieval-augmented generation (RAG) frameworks~\citep{malviya2024evidence, rothermel-etal-2024-infact, chern2023factool, xie2024improvingmodelfactualityfinegrained}. Many systems extend fact verification to large text corpora~\citep{schuster-etal-2022-stretching} by retrieving relevant passages~\citep{khattab2020colbertefficienteffectivepassage} and using language models to assess whether claims align with the retrieved content.

A wide range of Wikipedia-based fact verification datasets has been developed, including FEVER~\citep{thorne-etal-2018-fact}, FEVEROUS~\citep{feverous}, TabFact~\citep{chen2020tabfactlargescaledatasettablebased}, HOVER~\citep{jiang-etal-2020-hover}, WikiFactCheck-English~\citep{sathe-etal-2020-automated}, VitaminC~\citep{schuster-etal-2021-get}, EX-FEVER~\citep{ma-etal-2024-ex}, and AveriTeC~\citep{schlichtkrull-etal-2024-automated}. These datasets typically create the \refutes~class by synthetically modifying true statements, whereas our dataset captures contradictions naturally present in the corpus. WikiContradict~\citep{hou2024wikicontradict} also targets real contradictions but relies on inconsistency tags added by Wikipedia editors. Our analysis shows that many tagged cases have since been resolved, reducing the accuracy of those labels. Moreover, WikiContradict does not explicitly include a corpus-level \supports~class--- facts that are extensively checked to be free of corpus-level inconsistencies. WikiContradiction~\cite{hsu2021wikicontradiction} focuses on contradictions within a single article, whereas our \dataset extends the scope to contradictions across the entire corpus. This corpus-level setting introduces the additional challenge of searching for and aggregating evidence across multiple articles.

\paragraph{Claim Decomposition.}
The \task task requires decomposing the corpus into smaller, self-contained facts. Prior work has examined claim extraction and decomposition within fact verification systems~\citep{hu2024decompositiondilemmasdoesclaim, wührl2024selfadaptiveparaphrasingpreferencelearning, min2023factscore, song2024veriscore, cattan2024localizingfactualinconsistenciesattributable, gunjal-durrett-2024-molecular, pham-etal-2025-verify}.

%% file: 3_task.tex
\section{Corpus-Level Inconsistency Detection (\task)}\label{sec:task}

We define \task as a binary classification task over atomic facts. An atomic fact~\citep{min-etal-2023-factscore} is a short, self-contained statement that conveys a single piece of information and can be verified independently~\citep{semnani-etal-2023-wikichat, gunjal-durrett-2024-molecular}. An atomic fact from a corpus is \emph{corpus-level inconsistent} if there exists at least one other piece of information within the corpus that contradicts it; otherwise, it is consistent.

Formally, consider a corpus of documents $C = {D_1, D_2, \ldots, D_n}$. Let $f$ be an atomic fact extracted from some document $D_i \in C$. The objective is to determine whether there exists a subset of documents $E \subseteq C$ containing evidence that contradicts $f$. We define the function $\text{\task}(C, f) \mapsto \{\text{True}, \text{False}\}$ as:
\begin{equation*}
   \small
   \text{\task}(C, f) = \begin{cases}
      \text{True},    & \text{if } \exists E \subseteq C \text{ such that} \\
                     & \text{NLI}(E, f) = \refutes \\[0.5ex]
      \text{False},   & \text{otherwise} \\
   \end{cases}
\end{equation*}
where
\begin{equation*}
    \small
    \text{NLI}(E, f) \in \left\{
    \begin{array}{l}
        \supports, \refutes,\\
        \nei
    \end{array}
    \right\}
\end{equation*}
\noindent denotes the standard three-way Natural Language Inference task~\citep{bowman-etal-2015-large, condoravdi-etal-2003-entailment}.

\begin{figure*}[t!]
   \centering
   \includegraphics[width=0.9\linewidth]{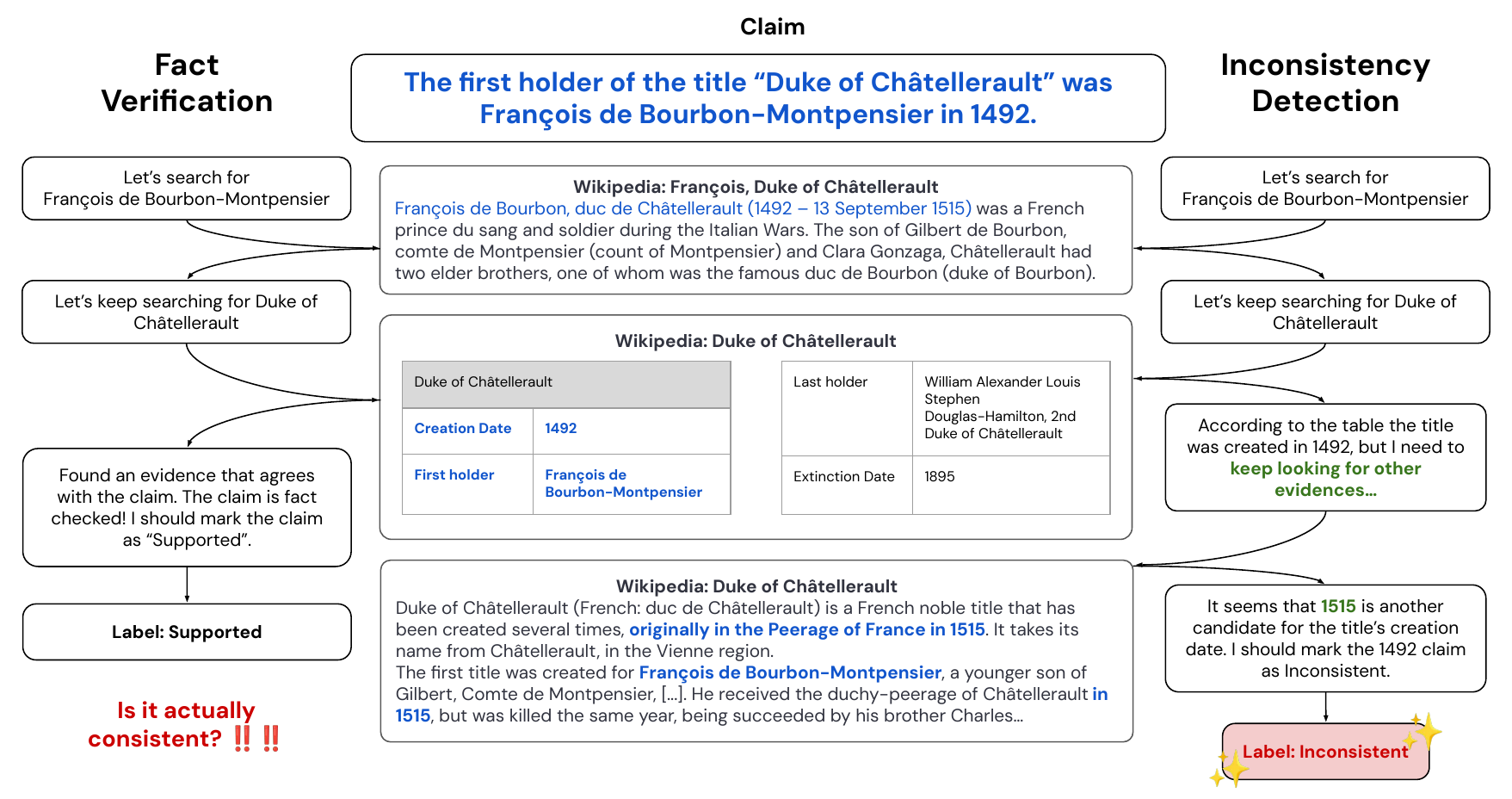}
   \caption{\textbf{An example from the \feverous dataset illustrating the difference between fact verification and inconsistency detection.} The claim is shortened for brevity. François de Bourbon-Montpensier was born in 1492 and received the title ``duchy-peerage of Châtellerault'' in 1515. However, the Wikipedia table ``Duke of Châtellerault'' incorrectly states that the title was created 23 years earlier. In fact verification, the corpus is assumed to be internally consistent, so the search may stop after finding one supporting piece of evidence. In inconsistency detection, the search continues to identify \emph{any} contradictory evidence within the corpus.}
   \label{fig:task}
\end{figure*}

\task is closely related to fact verification~\citep{thorne-etal-2018-fever, aly-etal-2021-fact} but differs in a critical assumption. Fact verification typically presumes that the corpus is internally consistent; the goal is therefore to find \emph{any} evidence supporting or refuting a given claim, i.e., $\exists E$ such that $\text{NLI}(E, f) \in {\supports, \refutes}$. In contrast, \task requires either identifying at least one piece of refuting evidence or \emph{exhaustively} verifying that no contradictory evidence exists anywhere in the corpus. Figure~\ref{fig:task} illustrates this distinction with an example from \feverous.

%% file: 4_system.tex
\section{\system: A Human-in-the-Loop Assistant for Corpus-Level Inconsistency Detection}

The \task task involves two primary subtasks:

\begin{enumerate}
\item \textbf{Research:} Gathering a comprehensive set of relevant evidence from a large corpus, as exhaustive manual checking is infeasible.
\item \textbf{Verification:} Determining whether any retrieved evidence contradicts the given fact.
\end{enumerate}

While humans generally perform well at verifying inconsistencies, our preliminary studies suggest they struggle to efficiently locate relevant pages that may contradict a given fact. To leverage the strengths of both humans and machines, we propose \system, an agent based on the ReAct architecture~\citep{yao2022react}. In this framework, research and verification steps are interleaved, allowing insights gained during verification to guide subsequent retrieval. This iterative process improves the agent's ability to uncover inconsistencies.

In our experiments, we found that simply presenting retrieval results can confuse users unfamiliar with the domain of the claim. Determining consistency often hinges on nuanced understandings of entities and concepts mentioned in the evidence. We therefore introduce two auxiliary actions to the research subtask and incorporate their outcome into the agent's outputs:

\begin{enumerate}
\item \action{clarify}:
Request clarifications to disambiguate entities.
To distinguish similarly named entities, the agent identifies ambiguities in the given fact and retrieved evidence, gathers additional context, and produces concise summaries highlighting key differences.
\item \action{explain}: Request explanations of specialized terminology. When encountering unfamiliar concepts (e.g., ``tie-break rules in tennis''), the agent queries an LLM for a brief, accessible explanation.
\end{enumerate}

This structure enables more targeted evidence collection, especially for complex or nuanced claims. Illustrative examples of the benefit of such retrieval appear in Appendix~\ref{sec:dataset_examples}. Implementation details are provided in Appendix~\ref{sec:tools_implementation}.

\system employs in-context learning with an LLM to assess whether retrieved evidence contradicts the given fact. Preliminary evaluations indicate that current LLMs alone do not reliably verify inconsistencies at high accuracy. Therefore, we design \system to output an inconsistency score in the range [0, 1] to quantify confidence and help users prioritize high-confidence candidates for inspection.

\subsection{User Study}
We evaluated the effectiveness of \system in helping users efficiently explore potential inconsistencies by conducting a user study with eight experienced Wikipedia editors (median number of edits: 2{,}124).

We integrated \system into a browser extension that highlights potentially inconsistent claims encountered during Wikipedia browsing and editing (Figure~\ref{fig:browser_extension}). The extension analyzes the current page in the background and, when a potential inconsistency is detected, highlights the claim and provides a tooltip with explanations and links to supporting evidence.

For each editor, we selected two Wikipedia articles from a pool of 10 that we had manually verified to contain multiple inconsistencies with the rest of Wikipedia. Each participant completed two 30-minute tasks in randomized order: (1) identifying inconsistencies in one article using our extension without external search, and (2) identifying inconsistencies in a different article without the extension, using any external tools they wish to use (including search engines and LLM chatbots). In both tasks, participants documented all inconsistencies they found within the assigned article.

Participants identified an average of 64.7\% more inconsistencies per hour when using \system.

\begin{figure}[ht!]
\centering
\includegraphics[width=0.48\textwidth]{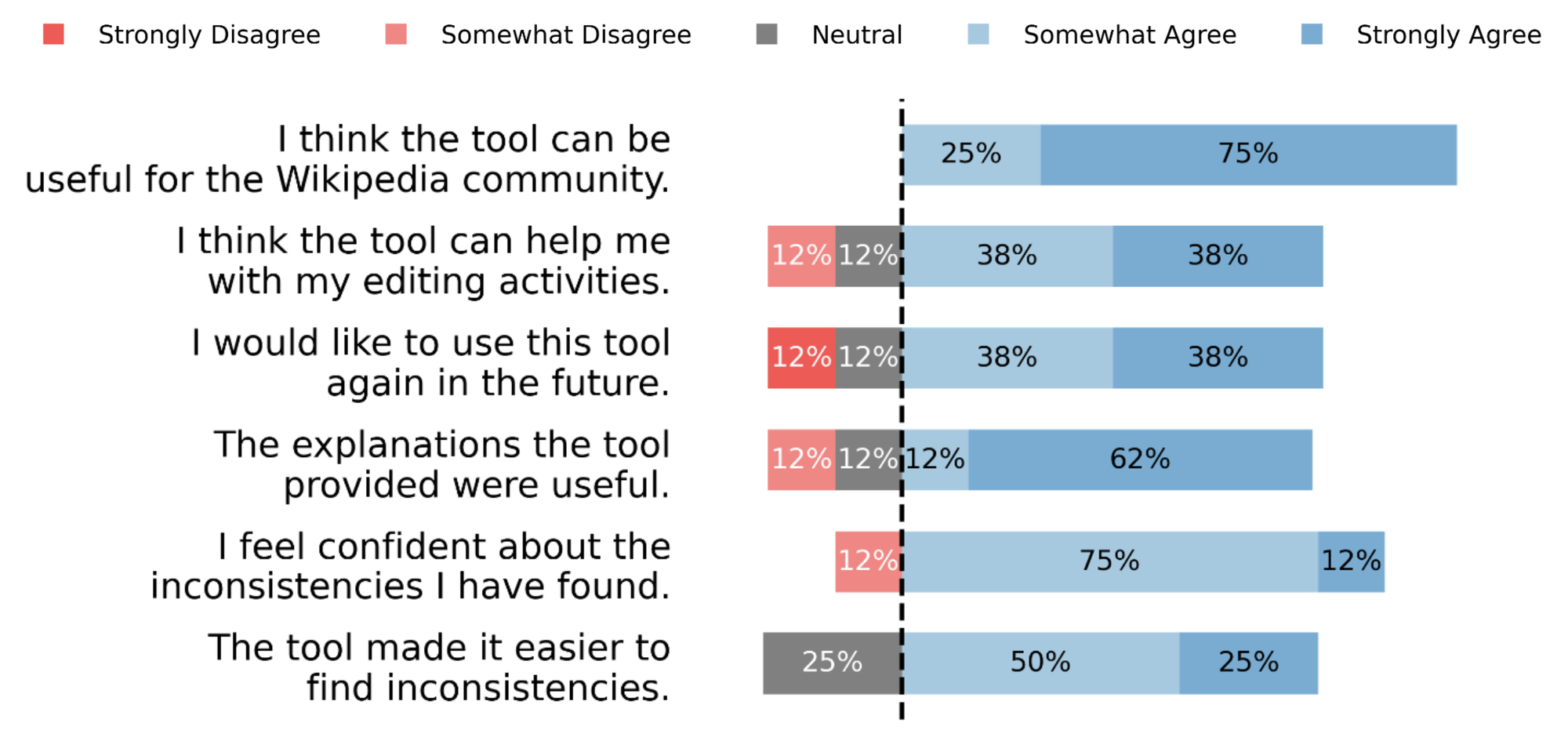}
\caption{\small Survey results on the perceived usefulness of our tool ($n=8$). Responses were collected using a 5-point Likert scale.}
\label{fig:usefulness_chart}
\end{figure}

After the tasks, we collected feedback on perceived usefulness. Participants rated their agreement with several statements on a 5-point Likert scale (\textit{Strongly Disagree} to \textit{Strongly Agree}); Figure~\ref{fig:usefulness_chart} shows the response distribution. Editors particularly valued the tool's ability to surface contradictions across article boundaries, information that typically requires extensive manual cross-referencing. Additionally, 87.5\% of participants reported increased confidence in identifying inconsistencies when using \system. These results suggest that AI-assisted inconsistency detection can effectively augment human curation.

Additional details and open-ended responses are provided in Appendix~\ref{sec:user_study}.

\section{Inconsistency Rates in the English Wikipedia and NLP Datasets}
\label{sec:implications}

We find that {\bf at least 3.3\% of Wikipedia facts are inconsistent.}
We establish a statistical lower bound on inconsistencies in the November 1, 2024 Wikipedia dump. Applying \system to 700 atomic facts uniformly sampled from Wikipedia articles, we identified 44 potentially inconsistent facts, of which 23 were manually confirmed inconsistent. With 99\% confidence, we estimate that approximately $3.3\% \pm 1.7\% [1.6\%, 5.0\%]$ of all facts in Wikipedia contradict other information in the corpus. This is a lower bound, as \system may miss inconsistencies (see Appendix~\ref{sec:inconsistency-calculation} for further details).
Extrapolated to the entire encyclopedia, this corresponds to between 37.6 million and 121.9 million inconsistent facts,\footnote{\url{https://en.wikipedia.org/wiki/Wikipedia:Size_of_Wikipedia}} underscoring the need for systematic inconsistency detection.

\textbf{Inconsistency rates vary across article categories.}
We further analyze frequency by mapping our uniform sample to Wikipedia article categories. Reliability varies substantially across domains, with narrative-heavy subjects particularly prone to inconsistencies. Articles in the “history” category exhibit the highest inconsistency rate (17.7\%), followed by Everyday Life (16.9\%) and Society \& Social Sciences (14.3\%) (Figure~\ref{fig:category_inconsistency}). The most common error type in history articles is numerical discrepancy. By contrast, categories requiring precise technical knowledge and quantifiable information---such as Mathematics (5.6\%) and Technology (9.4\%)---show markedly lower rates. See Appendix~\ref{section:inconsistency-in-categories} for more details.

\textbf{In the AmbigQA dataset~\citep{min-etal-2020-ambigqa}, we find that $4.0\% \pm 1.1\%$ of examples contradict information elsewhere in the corresponding Wikipedia dump}, reflecting underlying inconsistencies in Wikipedia.
This finding is significant given that AmbigQA is designed to have unique answers at the corpus level. For our investigation, we converted its question-answer pairs into declarative facts and applied the same methodology as before to assess inconsistency.

\textbf{Applying the same analysis to \feverous~\cite{aly-etal-2021-fact}, we find that $7.3\% \pm 0.5\%$ of claims labeled as \supports~are involved in corpus-level inconsistencies}: their verification outcome depends on which Wikipedia article is chosen as evidence and could have been labeled as \refutes~instead. This challenges the foundational assumption in fact verification that the corpus provides a consistent source of truth. 

%% file: 5_dataset.tex
\section{\dataset: A Dataset for Corpus-Level Inconsistency Detection}
\label{sec:dataset}

With the help of \system, we create the \dataset dataset, consisting of 955 atomic facts drawn from Wikipedia, each manually labeled as either consistent or inconsistent with the corpus. Whereas prior fact verification datasets often rely on synthetic contradictions that fail to capture real-world nuance, \dataset contains \emph{real, previously unknown} inconsistencies in Wikipedia.

For inconsistent facts, we provide manually verified evidence documents demonstrating the contradiction, detailed reasoning explaining the inconsistency, and a categorization of inconsistency type. For consistent facts, we provide up to 40 evidence passages from Wikipedia that were reviewed during annotation. Note that inconsistent labels represent a gold standard backed by concrete contradictory evidence, whereas consistent labels represent strong verification as exhaustively proving the absence of contradictions across a large corpus is infeasible.

The dataset highlights nuanced challenges such as implicit contradictions, temporal conflicts, and divergent interpretations that might otherwise go undetected. It covers diverse topics including people, history, geography, and science (Figure~\ref{fig:topic_distribution}), ensuring broad applicability across domains. Figure~\ref{tab:dataset_examples} shows representative examples from \dataset, illustrating the need for multi-hop reasoning, numerical calculations, nuanced context interpretation, entity disambiguation, and domain expertise.

\begin{table}[ht!]
   \centering
   \small
   \begin{tabular}{@{}lccc@{}}
      \toprule
      \textbf{Split} & \textbf{Inconsistent} & \textbf{Consistent} & \textbf{Total} \\
      \midrule
      Validation & 135 (28.3\%) & 342 (71.7\%) & 477 \\
      Test       & 196 (41.0\%) & 282 (59.0\%) & 478 \\
      \midrule
      \textbf{Total} & \textbf{331 (34.7\%)} & \textbf{624 (65.3\%)} & \textbf{955} \\
      \bottomrule
   \end{tabular}
   \caption{Distribution of consistent and inconsistent facts in \dataset.}
   \label{tab:dataset_statistics}
\vspace{-1em}
\end{table}

\subsection{Dataset Construction}

Constructing a corpus-level inconsistency dataset poses significant challenges. Given the rarity of inconsistencies, how can we efficiently identify sufficient and representative examples? Moreover, accurate annotation is difficult for both humans and machines. To address these challenges, we adopt a human-in-the-loop approach with \system. Figure~\ref{fig:dataset_construction} in Appendix~\ref{sec:annotation} summarizes the overall process.

\paragraph{Knowledge Corpus.} Because Wikipedia changes frequently, we use a frozen snapshot from November 1, 2024 for dataset construction and experiments to ensure reproducibility.

\paragraph{Selection of Facts for the Dataset.} We select facts through a three-step procedure:

\begin{enumerate}
\item \textbf{Sampling Popular Wikipedia Pages.} To ensure broad domain coverage, we sample from Wikipedia's Level 5 Vital Articles. These 50{,}000 articles are actively maintained by WikiProject Vital Articles and represent diverse topics and quality levels.\footnote{\url{https://en.wikipedia.org/wiki/Wikipedia:Content_assessment}} We extract text blocks delimited by newlines, filtering out passages that are too short (<100 characters) or too verbose (>320 characters) to maintain focused, verifiable content. From this filtered pool, we randomly sample 10{,}000 blocks while preserving the original category distribution.

\item \textbf{Fact Extraction.} Following prior work~\citep{semnani-etal-2023-wikichat}, we use GPT-4o (prompt in Figure~\ref{lst:claim_extraction}) to split each text block into atomic facts, yielding 89{,}300 atomic facts.

\item \textbf{Increasing the Proportion of Potentially Inconsistent Facts.} To obtain a relatively balanced dataset under high annotation cost, we prioritize facts more likely to be inconsistent. We apply a simple retrieval and LLM-based filtering method with high recall (Appendix~\ref{sec:weak_baseline}). Facts for which no relevant contradictory information is retrieved are filtered out, reducing the candidate set to 1{,}880 facts.
\end{enumerate}

\paragraph{Human-in-the-loop Annotation.} Annotation is performed by the authors and a small group of high-quality crowdworkers recruited via Prolific~\citep{prolific}. Annotators first verify that extracted facts faithfully reflect their source paragraphs, reducing the candidate set to 955 facts.

\begin{table*}[ht!]
\centering
\small
\renewcommand{\arraystretch}{1.2}
\begin{tabular}{@{\hspace{0.5em}}lp{0.7\textwidth}r@{\hspace{0.5em}}}
    \toprule
    
    \textbf{Inconsistency Type} & \textbf{Description} & \textbf{\%} \\
    
    \midrule
    \rowcolor{actionblue!30}
    \textbf{Numerical} & Inconsistencies in numerical data, such as quantities, measurements, or percentages & \textbf{54.7} \\
    \rowcolor{actionblue!10}
    \quad Off-by-One Numerical & Small discrepancy involving a margin of one unit & 23.0 \\
    \rowcolor{actionblue!10}
    \quad Clear Numerical & Significant difference that \emph{cannot} be explained by a margin of one unit & 31.7 \\
    
    \rowcolor{actionblue!30}
    \textbf{Logical} & The claim and evidence directly or indirectly contradict each other & \textbf{17.5} \\
    \rowcolor{actionblue!10}
    \quad Direct Logical & Clear negation or alternative to a unique fact & 14.8 \\
    \rowcolor{actionblue!10}
    \quad Indirect Logical & Contradiction inferred or indirectly implied & 2.7 \\
    \rowcolor{actionblue!30}
    \textbf{Definition} & Different definitions or interpretations for the same term or concept & \textbf{10.6} \\
    \rowcolor{actionblue!30}
    \textbf{Temporal} & Inconsistencies in dates, durations, or event sequences & \textbf{7.9} \\
    \rowcolor{actionblue!30}
    \textbf{Named Entity} & Inconsistencies identifying specific entities (people, organizations, locations) & \textbf{6.0} \\
    \rowcolor{actionblue!30}
    \textbf{Categorical} & Differences in categorizing entities, objects, or concepts & \textbf{2.1} \\
    \rowcolor{actionblue!30}
    \textbf{Spatial} & Inconsistencies in spatial descriptions or geographical information & \textbf{1.2} \\
    \bottomrule
\end{tabular}
\caption{Breakdown of inconsistency types in \dataset validation and test sets (331 inconsistent facts).}
\label{tab:inconsistency_breakdown}
\end{table*}

An annotation interface presents the findings of \system to annotators: (1) relevant documents from the corpus, (2) clarifications on ambiguous entities (e.g., people with identical names) and unfamiliar concepts, and (3) two-sided reasoning with both consistent and inconsistent interpretations of the gathered evidence (Appendix~\ref{sec:report_generation}). Annotators review this information to determine consistency and provide reasoning with citations. For each fact labeled consistent, annotators reviewed an average of 21 potential evidence passages.

\paragraph{Final Dataset.} The annotation effort yields 955 facts, of which 34.7\% are inconsistent. Facts are evenly split between validation and test sets in \dataset, with consistent and inconsistent labels randomly distributed. Table~\ref{tab:dataset_statistics} summarizes the dataset statistics.

\subsection{Analysis of \dataset}

We analyze the dataset to understand the sources and types of inconsistencies that appear in Wikipedia. We categorize inconsistencies into seven types; Table~\ref{tab:inconsistency_breakdown} reports their proportions.

Numerical discrepancies constitute 54.7\% of inconsistencies. Of these, 42\% are off-by-one errors, often involving dates or years in historical contexts; the remainder are more substantial and varied. Logical contradictions account for 17.5\%, with a smaller fraction requiring inference or indirect reasoning. The remaining 27.8\% arise from differing definitions, temporal or spatial conflicts, entity disambiguation errors, and divergent categorizations.

%% file: 6_experiment.tex
\section{Evaluating Automatic Corpus-Level Inconsistency Detectors using \dataset}
\label{sec:experiment}
With the dataset annotated, we evaluate multiple automated systems for \task.

\subsection{Evaluated Systems}

\textbf{\system.} We evaluate \system directly against the human-corrected labels.

\textbf{Retrieve-and-Verify.} Following established fact-checking methodologies~\citep{thorne-etal-2018-fever}, this system separates retrieval and verification. First, relevant passages are retrieved via similarity search. Then, a verification model (a single LLM call) assesses the consistency of the fact against all retrieved evidence and outputs an inconsistency score in [0, 1]. Facts with scores above 0.5 are classified as inconsistent.

\textbf{NLI Pipeline.} This system also follows a retrieve-and-verify approach but evaluates each retrieved passage individually against the fact using an LLM-based Natural Language Inference (NLI) model. Each evidence-fact pair is classified as \refutes, \supports~or \nei. A fact is marked inconsistent if at least one passage is classified as a contradiction.

\subsection{Experiment Setup}
We experiment with GPT-4o (\texttt{gpt-4o-2024-11-20}), the 70B-parameter LLaMA-3.1~\cite{grattafiori2024llama}, and \texttt{o3-mini}~\cite{openai2025o3} as LLM backbones. For retrieval, we embed all Wikipedia passages, tables, and infoboxes using the mGTE embedding model~\cite{zhang-etal-2024-mgte}. Unless noted otherwise, all experiments use RankGPT~\citep{sun-etal-2023-chatgpt} for reranking after retrieval.

The \system agent is allotted 10 steps, with 15 passages retrieved per query. For the retrieve-and-verify and NLI pipeline systems, we retrieve 20 passages per query, yielding a comparable total number of evidence items across methods for fair comparison. Ablation studies on these hyperparameters are provided in Section~\ref{sec:ablations}. Further implementation details and prompts for each system are provided in Appendix~\ref{sec:system_implementations}.

\subsection{Evaluation Metrics}
A primary use case of \task systems is flagging potential inconsistencies for human review. False positives waste human effort, while false negatives miss true inconsistencies. Therefore, we report the Area Under the Receiver Operating Characteristic curve (AUROC) as our main metric, alongside accuracy and F1.

For retrieve-and-verify and \system, we vary the inconsistency score threshold to compute ROC curves. For the NLI pipeline, we vary the number of contradictory passages required to classify a fact as inconsistent.

\subsection{Results}
Table~\ref{tab:results_main} reports performance using GPT-4o as the LLM backbone on the \dataset validation and test sets. \system achieves the best validation performance across all metrics. On the test set, \system outperforms other systems in Accuracy and AUROC by at least 0.3 and 2.1 points, respectively.

\begin{table}[ht!]
\centering
\small
\begin{tabular}{lccc}
    \toprule
    \textbf{System} & \textbf{Accuracy}  & \textbf{F1} & \textbf{AUROC} \\
    \midrule
    \multicolumn{4}{c}{\textbf{Validation set}} \\
    \system & \textbf{76.5} &  \textbf{67.4} & \textbf{80.9}  \\
    Retrieve-and-verify & 73.6 & 65.2 & 78.5 \\
    NLI-based pipeline & 74.0 & 66.5 & 78.4 \\
    \midrule
    \multicolumn{4}{c}{\textbf{Test set}} \\
    \system & \textbf{69.3} & 69.6 & \textbf{75.1} \\
    Retrieve and verify & 69.0 &  69.7 & 73.0 \\
    NLI pipeline & 67.0 & \textbf{70.2} & 72.2 \\
    \bottomrule
\end{tabular}

\caption{Overall performance of different systems using GPT-4o on the \dataset validation and test sets. The best score for each metric and split is shown in bold. For validation, we use a fixed threshold of 0.5. For test, thresholds are chosen to maximize validation F1: 0.6 for retrieve-and-verify, 0.5 for \system, and 1 contradictory passage for the NLI pipeline.}
\label{tab:results_main}
\end{table}

\subsection{Error Analysis}

All evaluated systems frequently conflate distinct entities that share the same name, leading to incorrect inconsistency flags.

A key challenge is context-dependent false positives: systems often detect discrepancies between a fact and retrieved evidence but misunderstand cases where those discrepancies are contextually acceptable. Below we detail cases where \system superficially and incorrectly flags inconsistencies due to limited contextual understanding:

\textbf{Numerical context.} Minor differences due to acceptable rounding or precision should not be flagged as inconsistent.

\textbf{Language context.} Articles sometimes include non-English terms whose translated forms differ for named entities; such translation variants should not be treated as inconsistencies. For example, the Japanese album title ``DoriMusu 1'' and its English equivalent ``Dreams 1'' are acceptable variants of the same entity name.

\textbf{Temporal context.} The system sometimes compares facts from different time periods and incorrectly flags inconsistencies when atomic facts lack explicit temporal qualifiers.

\begin{table}[ht!]
\centering
\small
\setlength{\tabcolsep}{3pt}
\begin{tabular}{lllll}
    \toprule
    \textbf{System} & \textbf{RR} & \textbf{Acc.} & \textbf{F1} & \textbf{AUROC} \\
    \midrule
    Retrieve+verify & \xmark & 71.9 & 62.8 & 76.5  \\
    ~ & \cmark & 73.6 \textcolor{ForestGreen}{\tiny(+1.7)}& 65.2 \textcolor{ForestGreen}{\tiny(+2.4)} & 78.5 \textcolor{ForestGreen}{\tiny(+2.0)} \\
    \system & \xmark & 74.6 & 64.9 & 78.1  \\
    ~ & \cmark & 76.5 \textcolor{ForestGreen}{\tiny(+1.9)}& 67.4 \textcolor{ForestGreen}{\tiny(+2.5)} & 80.9 \textcolor{ForestGreen}{\tiny(+2.8)}  \\
    \bottomrule
\end{tabular}
\caption{Ablation study showing the impact of reranking (RR) on system performance using GPT-4o on the \dataset validation set. Green values indicate improvements when reranking is applied. All systems use the same configurations as in Table~\ref{tab:results_main}.}
\label{tab:reranker_ablation}
\end{table}
    
\textbf{Perspective and belief context.} The system occasionally fails to distinguish differences in viewpoint, belief versus truth, or intention versus action. For example, it may incorrectly flag ``Alice believes Earth is flat'' as inconsistent with ``Bob believes Earth is round.''

\textbf{Legitimate variation in scholarly interpretation.} Apparent contradictions about historical events or scientific classifications may reflect evolving consensus rather than true inconsistencies.

\subsection{Ablation Studies}
\label{sec:ablations}

\begin{figure*}[ht!]
    \centering
    \begin{minipage}{0.48\textwidth}
        \centering
        \includegraphics[width=0.8\textwidth]{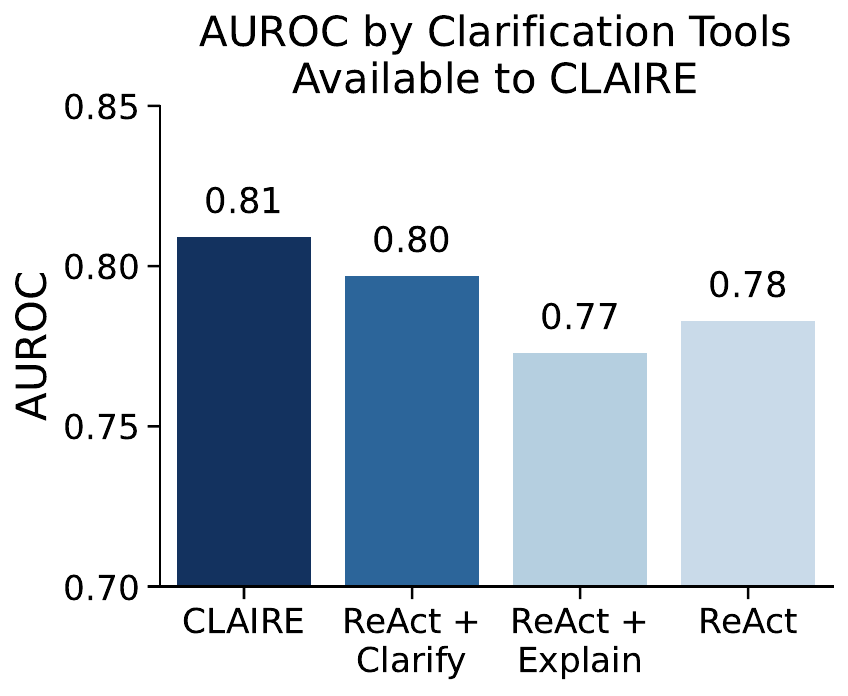}
        \caption{Ablation of tools available to the \system agent (GPT-4o) on the \dataset validation set. ReAct + Clarify and ReAct + Explain denote the ReAct agent with only one tool enabled.}
        \label{fig:ablation_react_tools}
    \end{minipage}
    \hfill
    \begin{minipage}{0.48\textwidth}
        \centering
        \includegraphics[width=0.8\textwidth]{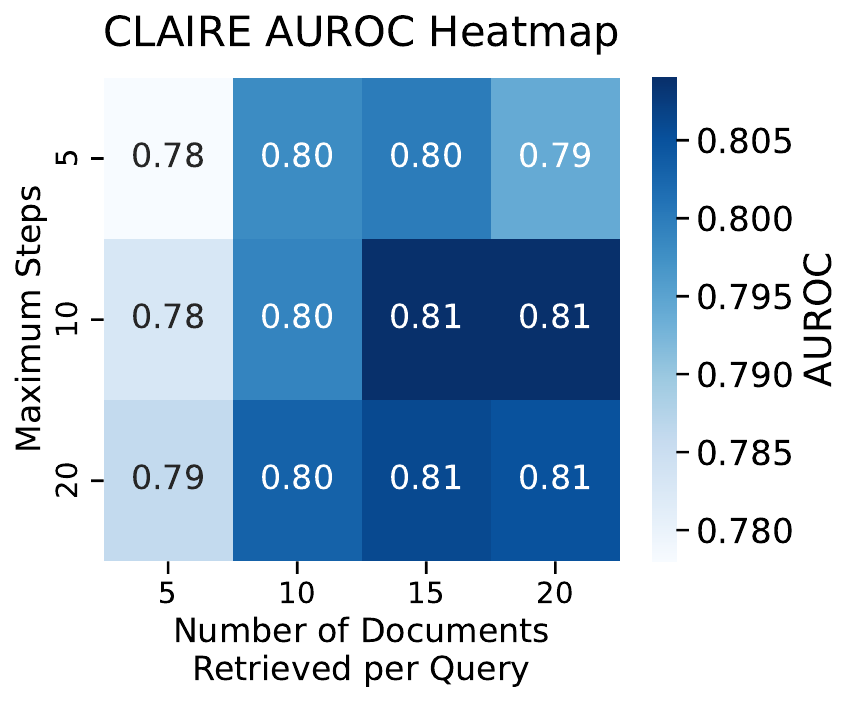}
        \caption{Ablation of retrieval hyperparameters for \system. Heatmap of AUROC as the number of steps and retrieved documents vary on the \dataset validation set.}
        \label{fig:ablation_react_heatmap}
    \end{minipage}
\end{figure*}

\textbf{Impact of tools available to the \system agent.}
Figure~\ref{fig:ablation_react_tools} shows that the agent achieves the highest F1 when using both \action{explain} and \action{clarify}.

\textbf{Impact of hyperparameters.} We vary (1) the number of thought-action-observation steps and (2) the number of documents returned per query. Figure~\ref{fig:ablation_react_heatmap} shows that \system generally performs better with more retrieved documents, but the effect is within 3\%.

\textbf{Impact of reranking in retrieval.} 
Embedding-based retrieval may miss deeper semantic relations. Adding a context-aware reranker prioritizes semantically relevant documents. As shown in Table~\ref{tab:reranker_ablation}, RankGPT reranking consistently improves all systems and metrics.

\begin{table}[ht!]
\centering
\small
\begin{tabular}{l@{\hspace{6pt}}l@{\hspace{4pt}}c@{\hspace{4pt}}c@{\hspace{4pt}}c}

    \toprule
    \textbf{System} & \textbf{Model} & \textbf{Accuracy} & \textbf{F1} & \textbf{AUROC} \\
    \midrule
    \multirow{3}{*}{\shortstack[l]{Retrieve\\and \\Verify}}  & GPT-4o & 73.6 &  65.2 & 78.5  \\
    & o3-mini & 75.7 &  65.7 & 77.0 \\
    & Llama-3.1-70B & 67.9 & 52.3 & 70.9 \\
    \midrule
   
    \multirow{3}{*}{\shortstack[l]{NLI \\Pipeline}} & GPT-4o & 74.0 & 66.5 & 78.4  \\
    & o3-mini & 65.4 & 59.5 & 77.0 \\
    & Llama-3.1-70B & 63.1 & 53.9 & 65.6 \\

    \midrule
    \multirow{3}{*}{\shortstack[l]{\system\\(ours)}}  & GPT-4o & \textbf{76.5} & \textbf{67.4} & \textbf{80.9} \\
    & o3-mini & 76.3  & 54.6 & 68.1 \\
    & Llama-3.1-70B & 69.0 & 43.9 & 69.5 \\
    \bottomrule
\end{tabular}
\caption{Ablation study of different LLMs on the \dataset validation set. The best score for each metric is shown in bold.}
\label{tab:ablation_model}
\end{table}

\textbf{Impact of the LLM used.}
We compare GPT-4o, o3-mini (medium reasoning), and Llama-3.1-70B on the validation set. As shown in Table~\ref{tab:ablation_model}, GPT-4o consistently achieves the highest scores. o3-mini is competitive, with generally higher precision; however, using the same prompt, it rarely outputs inconsistency scores in the intermediate range (0.1--0.9), instead clustering at extremes. Llama-3.1-70B underperforms relative to the other two.

%% file: 7_conclusion.tex
\section{Conclusion}
\label{sec:conclusion}

We introduce Corpus-Level Inconsistency Detection (\task), addressing the challenge of identifying contradictory information within large knowledge repositories. To tackle this problem, we present \system, an agent-based system that combines retrieval with LLM reasoning to detect and contextualize potential contradictions for human review.

We also release \dataset, a benchmark capturing real inconsistencies that synthetic datasets often miss. Our experiments show that, while retrieval and verification are challenging, \system enables humans to uncover substantially more inconsistencies. Applied to Wikipedia, this framework reveals that approximately 3.3\% of facts conflict with other information in the corpus, amounting to millions of contradictory statements across the encyclopedia.

These results demonstrate that corpus-level inconsistencies are a measurable phenomenon in large-scale knowledge corpora. Although automated systems still exhibit systematic errors, they can aid in maintaining knowledge consistency at scale. More broadly, this work suggests a virtuous cycle: LLMs help curate cleaner, more reliable corpora, which in turn improve both human knowledge access and the AI systems built on top of them.

%% file: 8_other.tex
\section*{Limitations}
This paper focuses exclusively on Wikipedia, the largest open text corpus. As a result, we do not explore other potentially valuable applications of corpus-level inconsistency detection, such as technical texts (e.g., academic, medical, or legal documents) or structured data sources like databases and knowledge graphs. We also leave detecting cross-lingual inconsistencies across different language versions of Wikipedia to future work.

\section*{Ethical Considerations}
We do not anticipate risks or ethical concerns arising from the publication of \dataset.

For crowdsourcing, we compensated annotators per task, with an overall rate of at least \$16 per hour. Participants in our user study were compensated at \$20 per hour. The study was approved by our institution's IRB, and participants provided informed consent. No personally identifiable information was collected during annotation or the user study. We release \dataset under a license compatible with Wikipedia's license.

Regarding computational resources, we used a CPU-based machine to serve the Wikipedia index and relied on commercial LLM APIs, making direct estimation of carbon footprint difficult. As a proxy, the total experimental cost did not exceed \$4,000.

\section*{Acknowledgment}
This work is supported in part by the Verdant Foundation, the Alfred P. Sloan Foundation, Microsoft Azure AI credits, and the NAIRR Pilot program.

%% file: 9_appendix.tex
\appendix

\section{Establishing Lower Bounds for Inconsistency Rates}
\label{sec:inconsistency-calculation}

To determine an appropriate sample size for estimating the proportion of inconsistencies in Wikipedia, we compute the minimum number of claims required to achieve a 99\% confidence level with a 5\% margin of error using the Cochran formula~\citep{cochran1953sampling}:
\begin{align*}
n &= \frac{z^2 \times p(1-p)}{E^2}  \\ &= \frac{2.576^2 \times 0.5 \times 0.5}{0.05^2} = 664
\end{align*}
where $z = 2.576$ corresponds to a 99\% confidence level, $p = 0.5$ assumes maximum variance (yielding the largest required sample size), and $E = 0.05$ is the desired margin of error. This calculation indicates that we must examine at least 664 claims. We apply the same statistical approach to determine sample sizes for our analyses of the AmbigQA and FEVEROUS datasets.

\subsection{Inconsistencies Per Wikipedia Article Category}
\label{section:inconsistency-in-categories}
\begin{figure}[ht!]
\centering
\includegraphics[width=\linewidth]{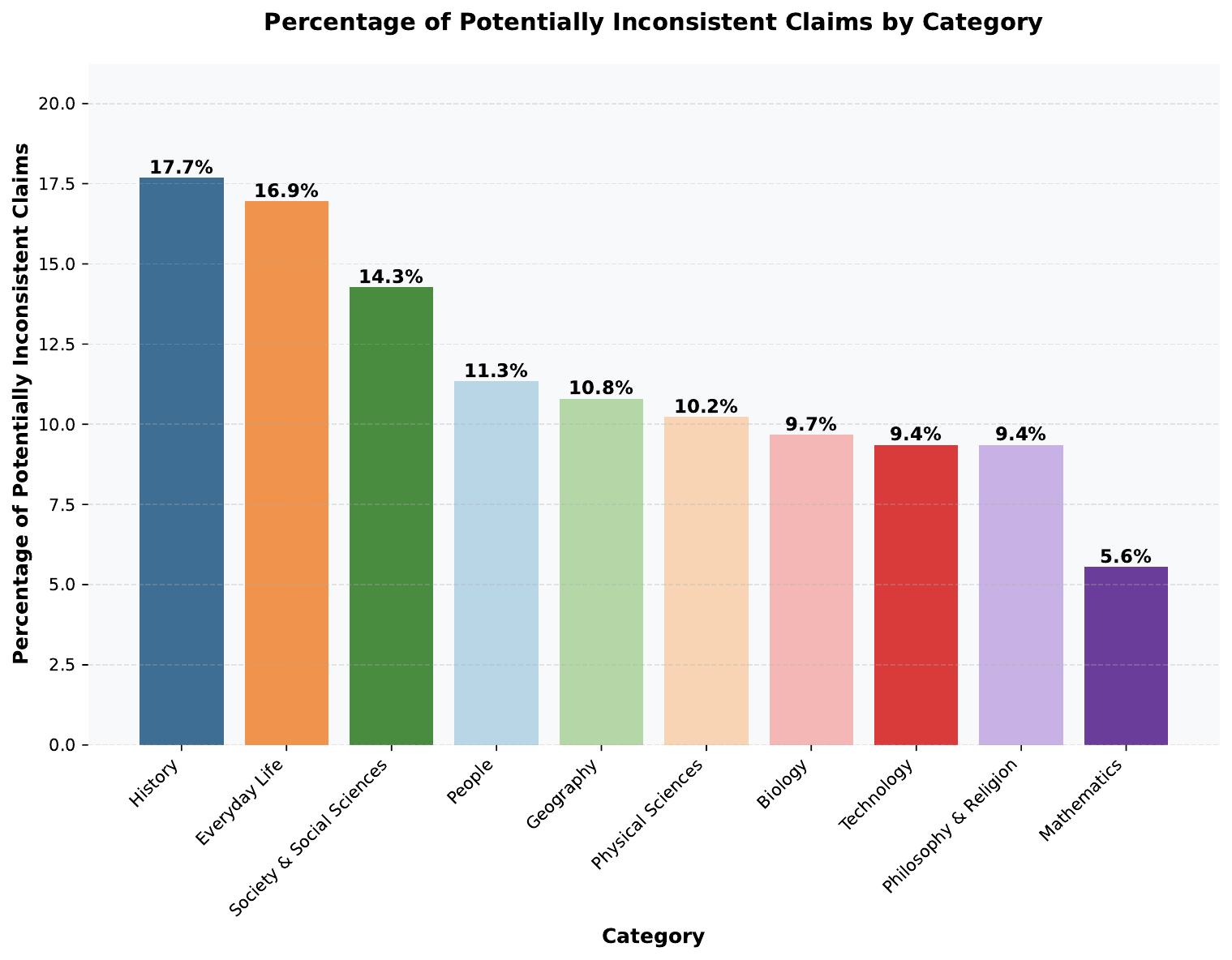}
\caption{Distribution of inconsistencies in Wikipedia across topics.}
\label{fig:category_inconsistency}
\end{figure}

The distribution of potentially inconsistent claims across Wikipedia categories reveals notable patterns (Figure~\ref{fig:category_inconsistency}). History exhibits the highest rate of inconsistencies (17.7\%), followed by Everyday Life (16.9\%) and Society \& Social Sciences (14.3\%). These trends are reflected in concrete cases from our analysis. For example, the claim that ``The Ottoman Empire first developed the technique of using explosive shells in naval warfare in 1640'' is inconsistent with historical records documenting earlier use by other naval powers. Similarly, the assertion that ``London's population doubled between 1800 and 1820'' oversimplifies gradual demographic change; empirical population estimates for that period do not support a doubling.

In contrast, categories requiring precise technical knowledge, such as Mathematics (5.6\%) and Technology (9.4\%), show markedly lower inconsistency rates, suggesting that factual precision is better maintained in domains with more quantifiable information. Overall, these results indicate that Wikipedia's reliability varies across knowledge domains, with narrative-heavy subjects being particularly susceptible to inconsistencies.

\section{More Details on the Annotation Process for \dataset}\label{sec:annotation}

\subsection{Annotation Tool}
\begin{figure*}[ht!]
\centering
\includegraphics[width=\linewidth]{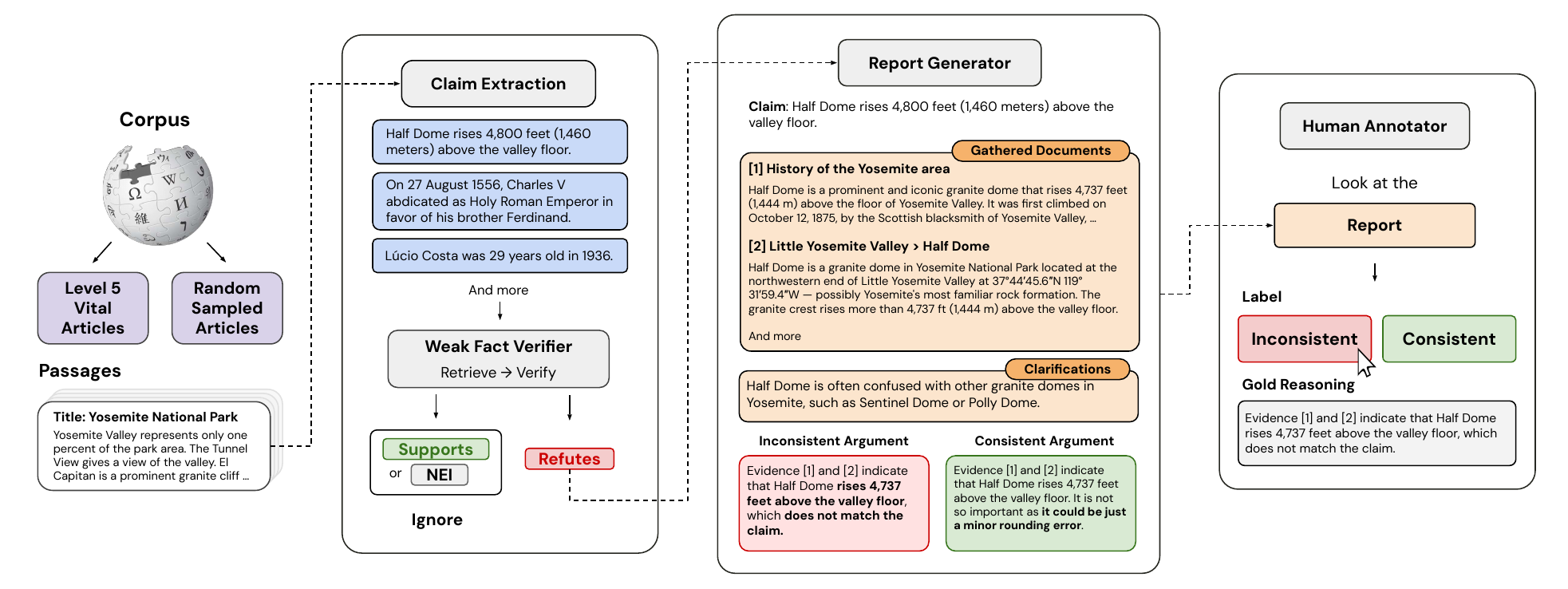}
\caption{Overview of the \dataset~construction process: diverse passage sampling from Wikipedia's Vital Articles, adversarial claims collection using GPT-4o and a weak baseline filter, and human verification with detailed evidence analysis. Randomly sampled articles are used to estimate the prevalence of inconsistencies in the entire English Wikipedia.}
\label{fig:dataset_construction}
\end{figure*}

Figure~\ref{fig:dataset_construction} provides an overview of the dataset construction process.

\paragraph{Filter to Balance Inconsistent Labels in the Dataset.}\label{sec:weak_baseline}
We implement a weak baseline to provide a permissive standard for inconsistency detection and filter out obviously consistent claims during \dataset construction. This baseline is a simplified version of the retrieve-and-verify system described in Section\ref{sec:experiment}, where the verifier outputs a binary inconsistency decision rather than a confidence score. We use GPT-4o mini~\citep{gpt4o} as the language model for these binary decisions.

\begin{figure*}[ht!]
\centering
\includegraphics[width=\textwidth]{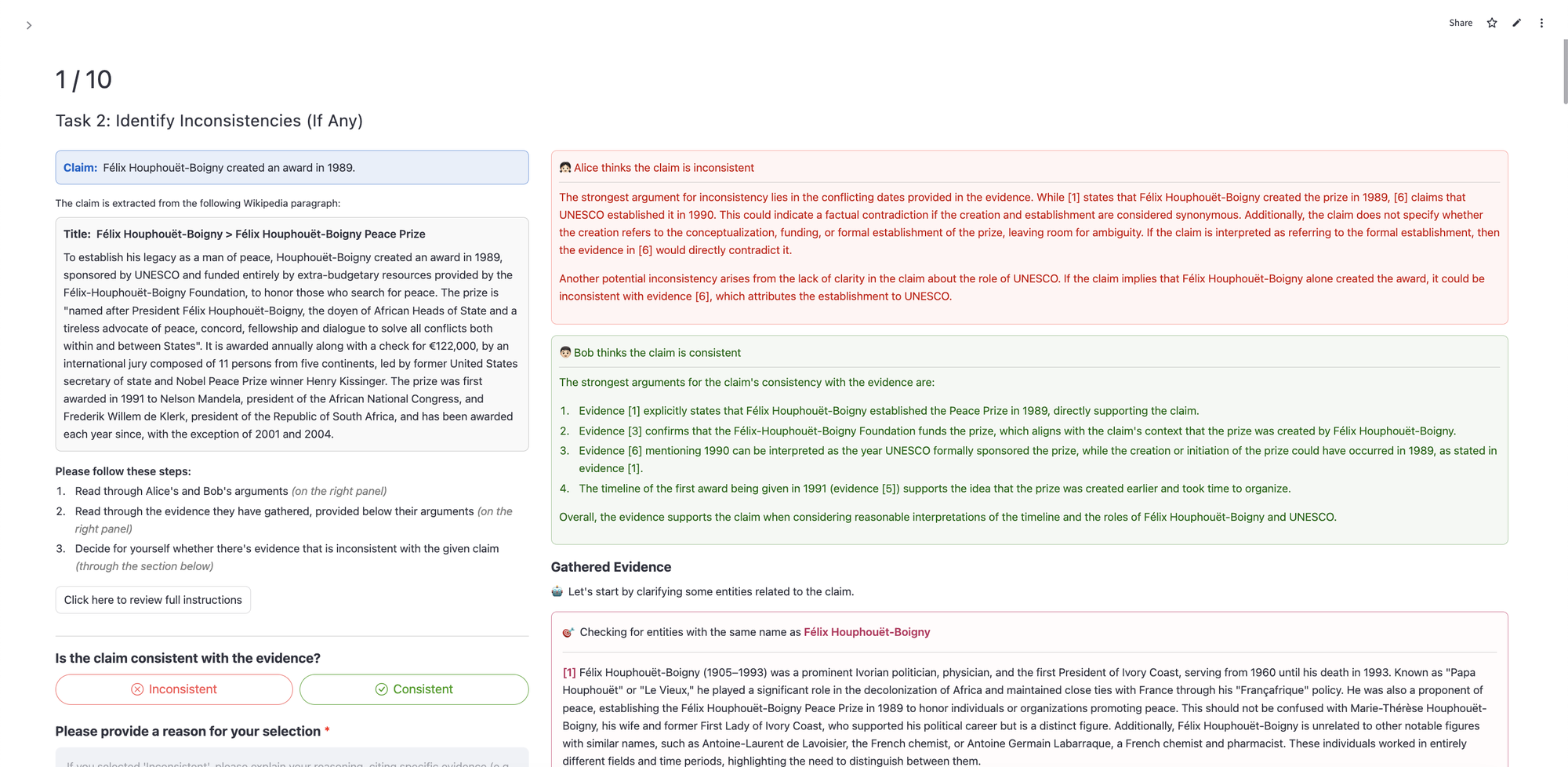}
\caption{Screenshot 2 of the annotation tool showing the main interface for claim verification and inconsistency labeling.}
\label{fig:annotation_tool_2}
\end{figure*}

\begin{figure*}[ht!]
\centering
\includegraphics[width=\textwidth]{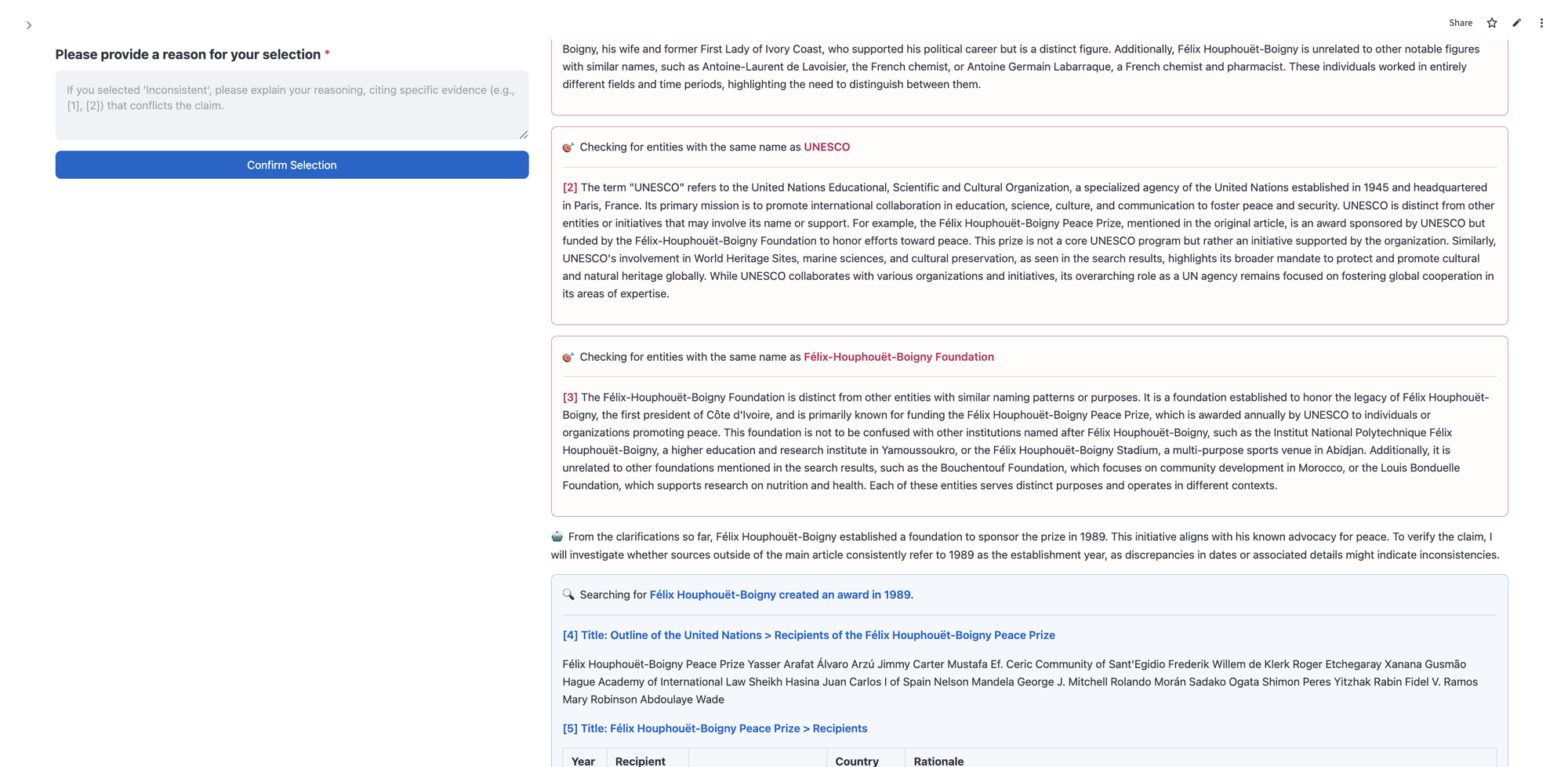}
\caption{Screenshot 3 of the annotation tool showing the main interface for claim verification and inconsistency labeling.}
\label{fig:annotation_tool_3}
\end{figure*}

\begin{figure*}[ht!]
\centering
\includegraphics[width=\textwidth]{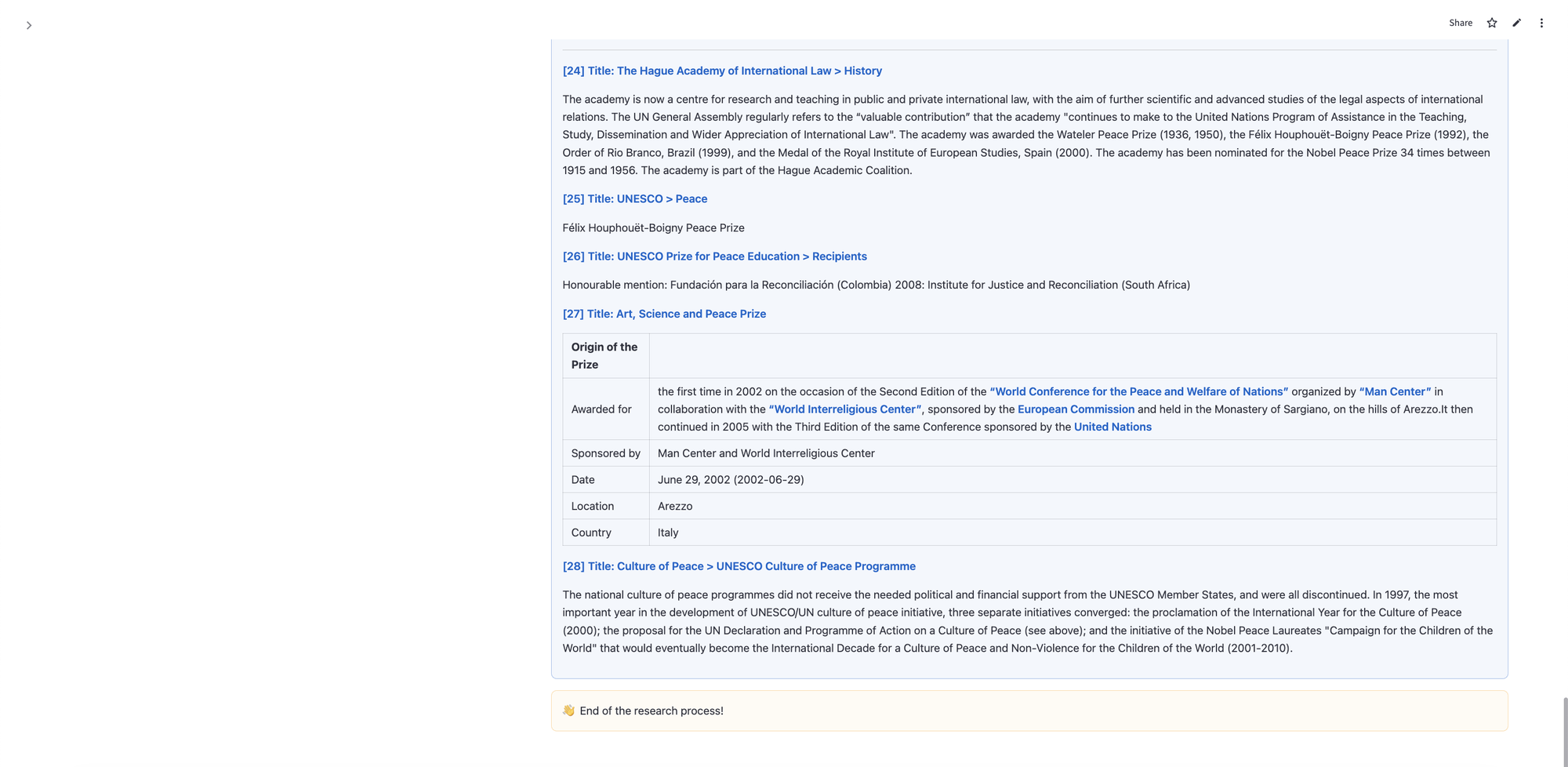}
\caption{Screenshot 4 of the annotation tool showing the main interface for claim verification and inconsistency labeling.}
\label{fig:annotation_tool_4}
\end{figure*}

\paragraph{Report Generation.}\label{sec:report_generation}
We develop a report generation system that produces a detailed report for each fact in the dataset. This system mirrors the setup in Section~\ref{sec:experiment} but replaces the verifier with a report generation stage. The report stage takes all retrieved evidence and the clarifications made by the tools agent and generates a two-sided analysis via two GPT-4o calls: one soliciting reasoning that the fact is inconsistent and another soliciting reasoning that it is consistent. This provides annotators with balanced information for final judgment. The final report shown to annotators includes both lines of reasoning and the agent's trace. See Figure~\ref{fig:annotation_tool_2} through~\ref{fig:annotation_tool_4} for illustrations.

\begin{figure*}[ht!]
\centering
\includegraphics[width=\textwidth]{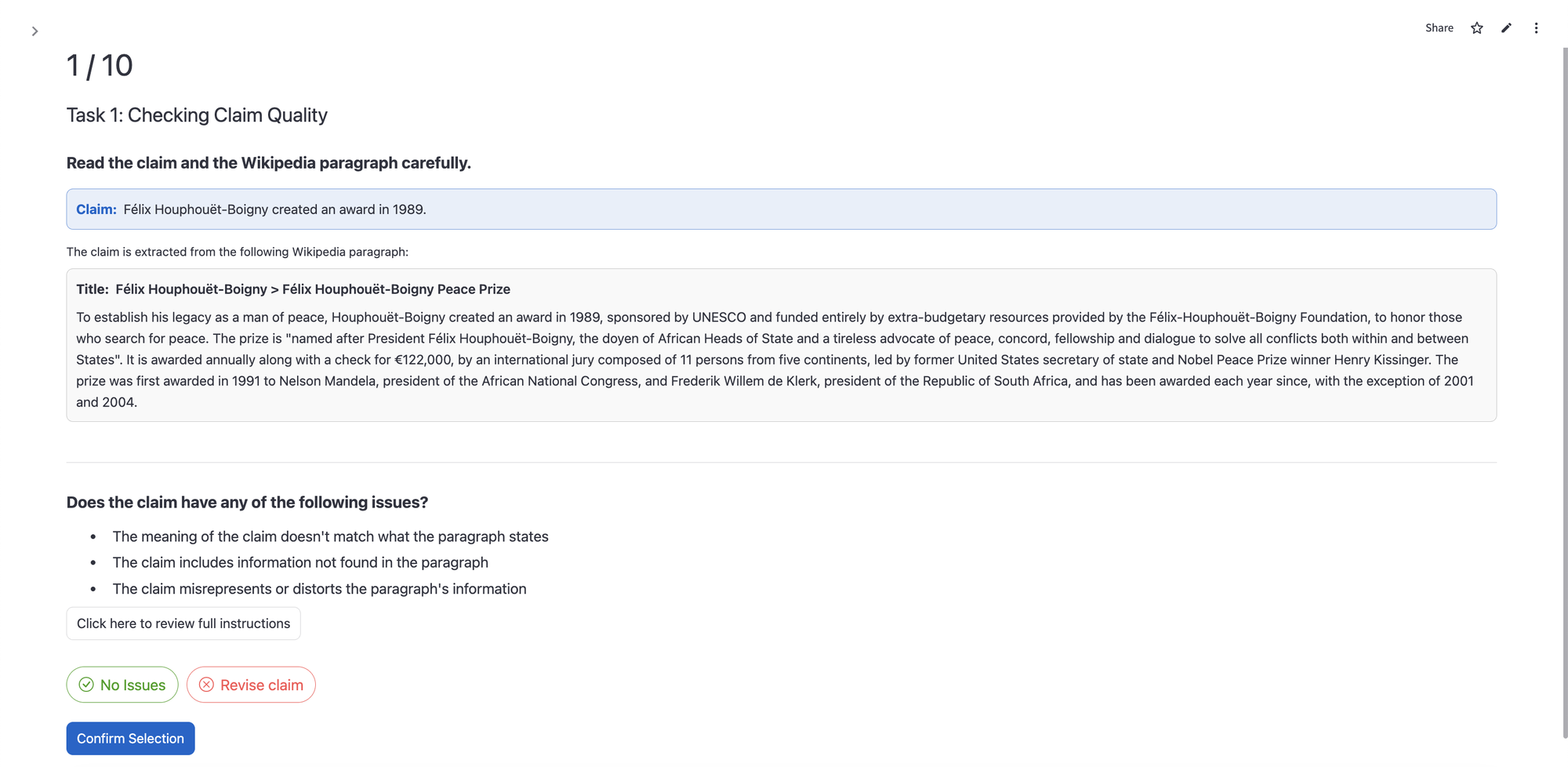}
\caption{Screenshot 1 of the annotation tool showing the main interface for claim verification and inconsistency labeling.}
\label{fig:annotation_tool_1}
\end{figure*}

\paragraph{Annotation Portal.} We built a web-based annotation platform to streamline the workflow. Annotators first check the extracted fact for any extraction issues, then review the detailed inconsistency analysis report, and finally assign a label of either ``consistent'' or ``inconsistent''. Screenshots of the interface are shown in Figure~\ref{fig:annotation_tool_1} through~\ref{fig:annotation_tool_4}.

\subsection{Annotators}
The dataset is annotated partly by the authors and partly by annotators recruited via Prolific~\cite{prolific}. To recruit external annotators, we conducted an initial qualification test using 10 randomly sampled facts from a subset previously labeled by the authors. Candidates were evaluated on both labeling accuracy and the quality of their written justifications. We selected the top 17 candidates who demonstrated strong analytical skills and high accuracy in identifying inconsistencies. Because inconsistency detection is nuanced, we required annotators to be native English speakers from the US or UK, hold a graduate degree (Masters or PhD), and maintain a Prolific approval rate of at least 95\%. The final pool of 17 annotators spent an average of 6.5 minutes evaluating each fact.

\subsection{Annotation Guidelines}
The task is inherently complex, which increases the risk of labeling errors. Determining whether a claim is inconsistent with a set of evidence is substantially more challenging than many standard annotation tasks. The definition of inconsistency can be context-dependent. For example, if a claim states a population of 4.8 million while the evidence reports 5 million, the case could be labeled ``inconsistent'' under exact matching or ``consistent'' if rounding is deemed acceptable. Likewise, comparing an imprecise expression such as ``a few years'' to a specific value is nontrivial, and the threshold for inconsistency is not always clear.

To mitigate ambiguity, we established explicit guidelines for such cases and instructed annotators to follow them closely.

\section{Dataset Details}

\subsection{Distribution of the Topics}\label{sec:dataset_analysis}
\begin{figure*}[ht!]
\centering
\includegraphics[width=0.6\linewidth]{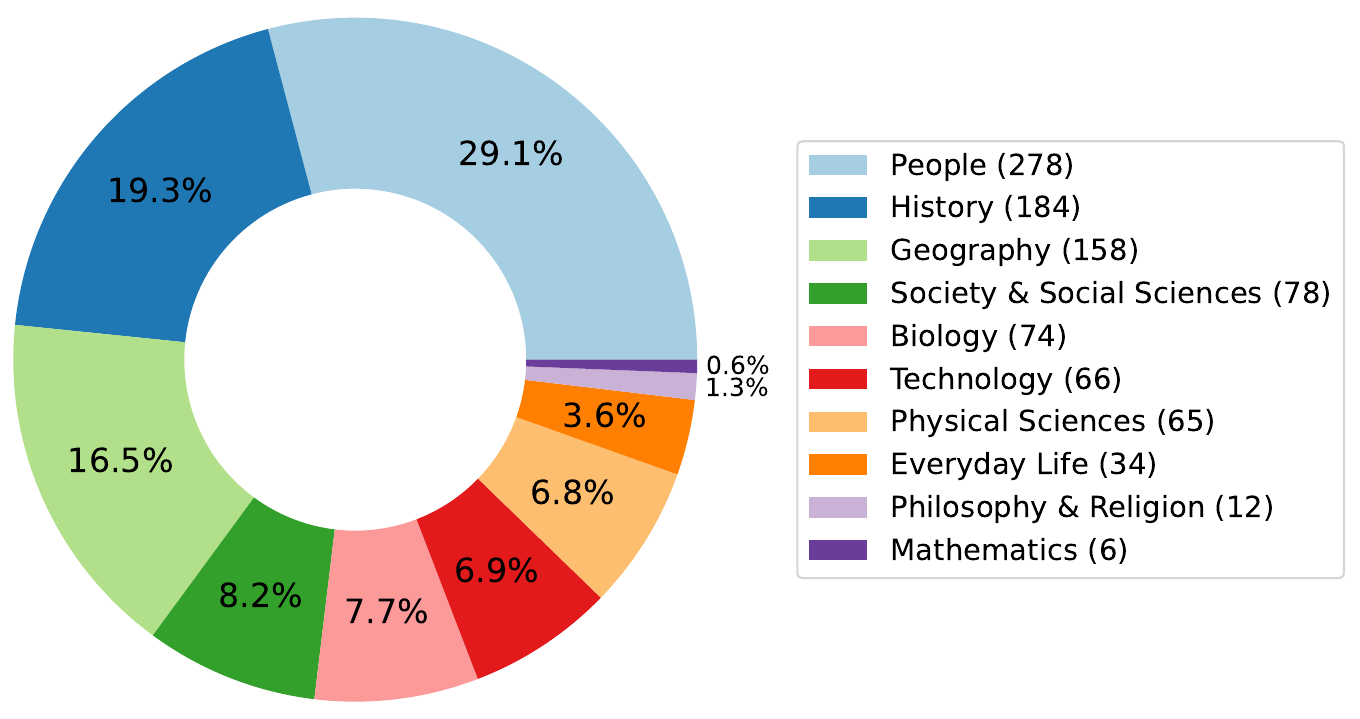}
\caption{Distribution of topics across the \dataset dataset, showing the diversity of knowledge domains covered.}
\label{fig:topic_distribution}
\end{figure*}

Figure~\ref{fig:topic_distribution} shows the distribution of topics covered by the facts.

\subsection{Dataset Examples}
\label{sec:dataset_examples}
\begin{figure*}[htbp]
\centering
\includegraphics[width=\linewidth]{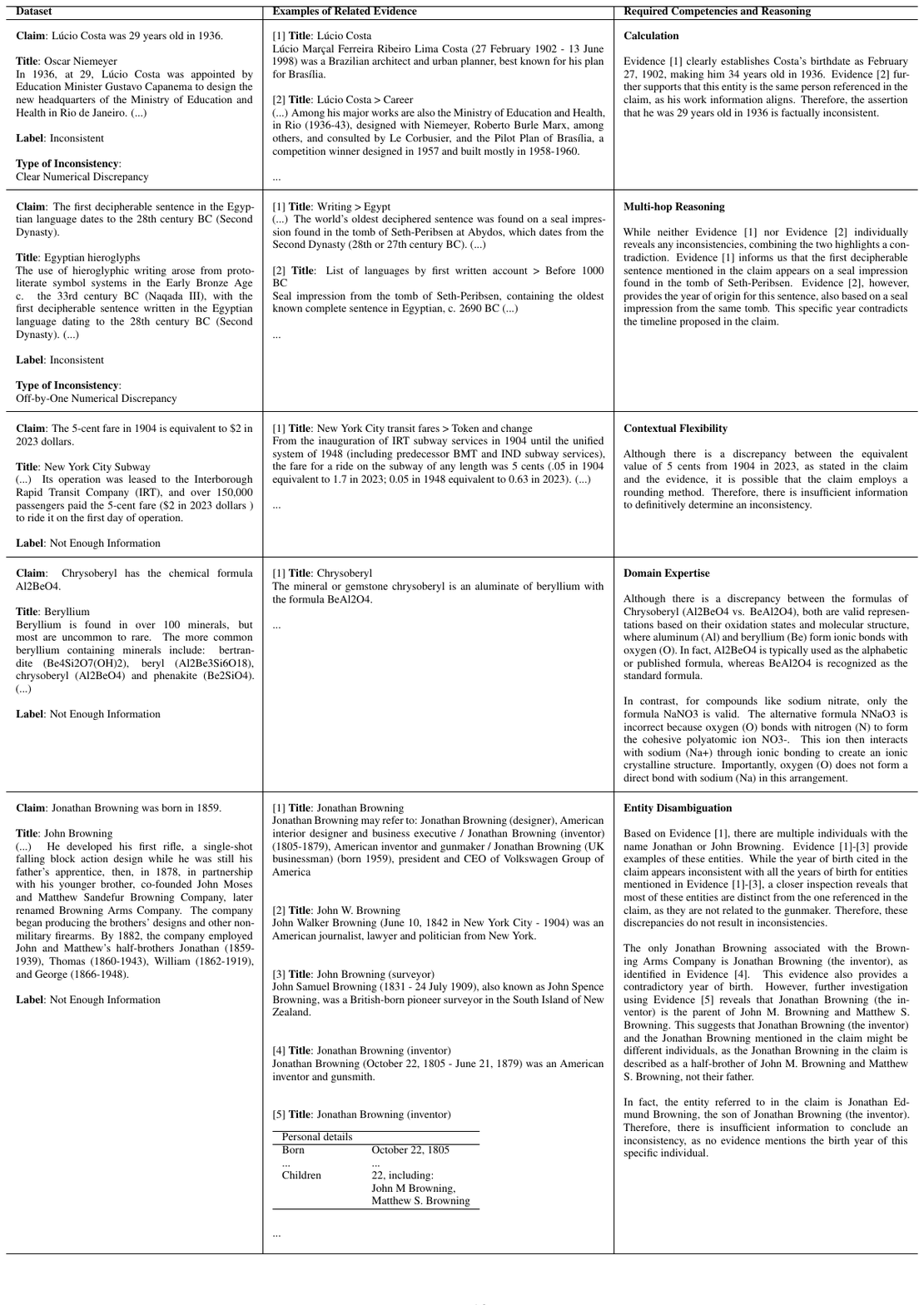}
\caption{Examples of claims with evidence from the dataset and required competencies.}
\label{tab:dataset_examples}
\end{figure*}

Figure~\ref{tab:dataset_examples} presents detailed examples of claims with accompanying evidence from the dataset.

\section{More Details on the User Study}
\label{sec:user_study}

\subsection{Browser Extension Implementation}

We implement the browser extension using JavaScript for the frontend and Python for the backend server. The frontend is a lightweight Chrome extension that injects content scripts to highlight potentially inconsistent claims on Wikipedia pages and provide a potential explanation for the inconsistency in the form of a side panel. For fact extraction, we use GPT-4o to parse Wikipedia page content into atomic claims, following prior work~\citep{semnani-etal-2023-wikichat,min-etal-2023-factscore}. The extension communicates with the backend via REST API endpoints. 

\begin{figure*}[ht!]
    \centering
    \includegraphics[width=\textwidth]{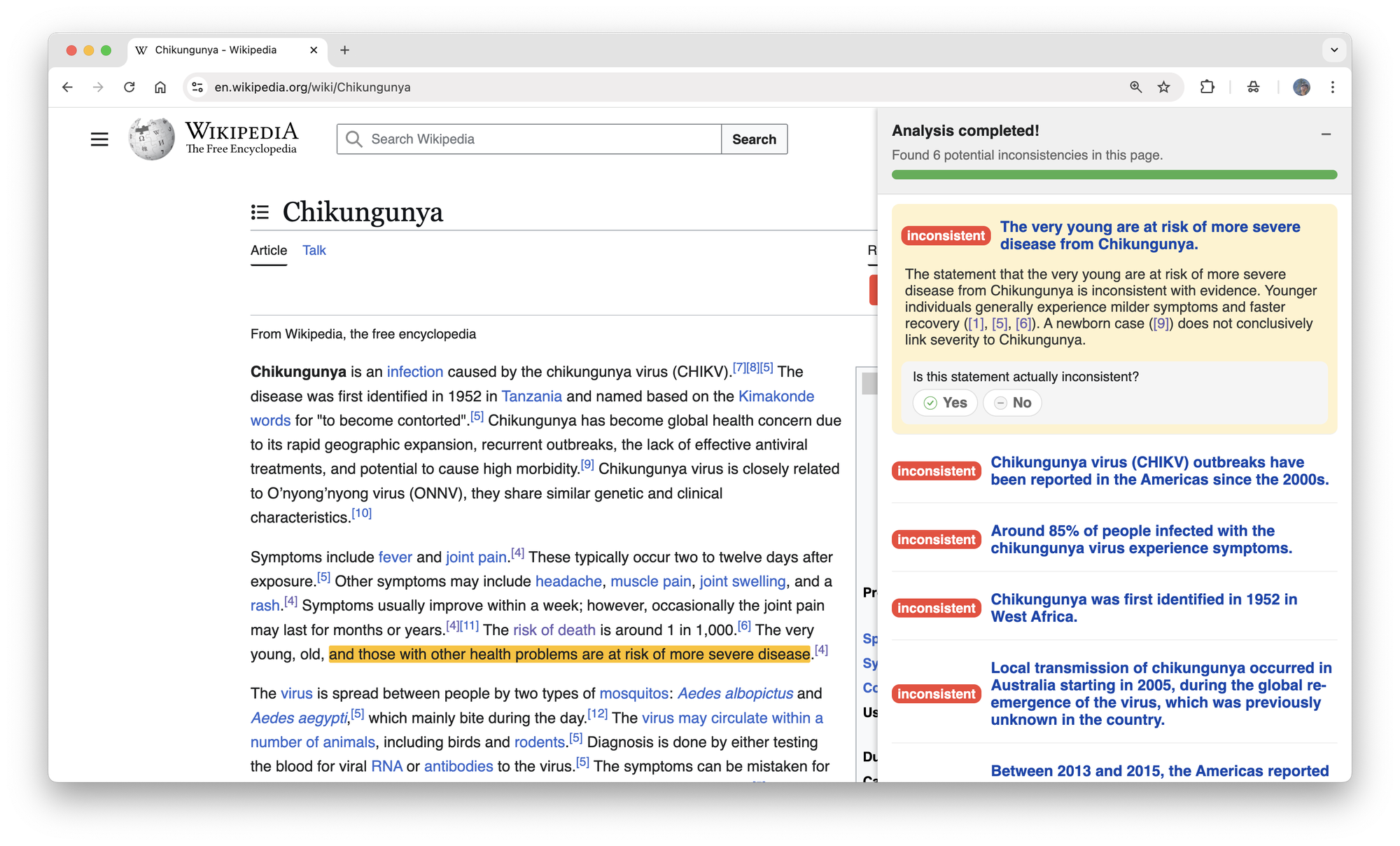}
    \caption{The browser extension is implemented as a button in Wikipedia which the user can click to check for inconsistencies. When a claim is flagged as potentially inconsistent, we highlight the claim on the page and show a side panel with a detailed explanation of the inconsistency and links to the evidence documents.}
    \label{fig:browser_extension}
\end{figure*}

\subsection{Recruitment Details}

We recruited participants through by posting a research participation call on Meta-Wiki\footnote{\url{https://meta.wikimedia.org}}.
All recruited editors had made at least 200 edits and been active for over one year. The study protocol was approved by our institution's Institutional Review Board (IRB), and all participants provided informed consent before beginning the study.

\subsection{Perceived Usefulness and Qualitative Feecback}

In addition to Likert-scale ratings, we collect open-ended feedback about participants' experiences finding inconsistencies with and without our tool. Editors report that they employ diverse manual strategies, such as cross-checking linked articles, search and keyword search, reviewing talk pages, and validating references, but find these approaches time-consuming and cognitively cumbersome. They further express that they value the tool's accessibility for novice editors and utility for verifying AI-generated content. The main concerns center on processing speed, false positive rates, and interface design. These insights reveal a key tradeoff: while our system significantly reduces the editorial burden, its effectiveness ultimately depends on balancing detection sensitivity with efficiency and usability. See Table~\ref{tab:user_feedback} for examples of user feedback.

\begin{table*}[htbp]
\centering
\begin{tabular}{p{0.95\linewidth}}
\toprule
\textbf{Positive Feedback} \\
\midrule
The fact that I can simply enable the extension and, within a few minutes, see at a glance 126 inconsistencies is impressive. I also appreciate that it directly links me to where each inconsistency occurs in the article and provides a brief summary. Additionally, I like the feature that allows me to validate these inconsistencies by either accepting or rejecting them. \\
\midrule
It gave a lot of potential statements and claims that could be inconsistent with other information. One of the harder parts without using the tool was to find relevant articles that would include overlapping information. \\
\midrule
The tool is instructive, directive and straight to the point. It identifies inconsistency without rigorous means. \\
\midrule
Very good at finding discrepancies. \\
\midrule
That each result has an option to give feedback if the statement is actually inconsistent. \\
\midrule
Really helpful automation, especially for editors that may not be a subject-matter expert. Like that it could be used to fact-check AI-generated content if/when it comes to Wikipedia. \\
\midrule
It clearly demarked which statements were potentially inconsistent and with what, plus it allows for feedback on if the statements are in fact inconsistent. \\
\midrule
I really like the intent of the tool! I feel like it holds a lot of promise to help editors fact-check and correct inaccurate articles more quickly. I also love the UI - it's simple, clear, and easy to interpret. I especially appreciated the explanations offered in the panel, which even linked to helpful sources! That part was really impressive. I was able to detect one very clear inconsistency using this tool, and I was able to identify and validate it extremely quickly because of the helpful explanation and sources that the tool provided. \\
\midrule
\textbf{Areas for Improvement} \\
\midrule
I didn't like how often the tool was wrong. Even if, on paper, it would be better to highlight claims that the tool is unsure is consistent, in reality I personally felt very annoyed every time it was wrong. \\
\midrule
It assumed that Georgian, Renaissance Revival, etc. architecture were mutually exclusive with Palladian architecture, and that all the American buildings listed as having been influenced by Palladian architecture were inconsistent due to the architectural style listed in those articles. \\
\midrule
UI/UX improvements mostly. Would be nice if it was a native feature that could be turned on in Wikipedia and did not require downloading onto your computer. \\
\midrule
There wasn't much to dislike! I think that the model's precision could be improved, but the experience itself was really great. \\
\bottomrule
\end{tabular}
\caption{Qualitative User Feedback on \system}
\label{tab:user_feedback}
\end{table*}

\section{Implementation Details}
\label{sec:system_implementations}
We accessed OpenAI models via Azure OpenAI and accessed LLaMA models through Azure AI Services. For all experiments, greedy decoding (i.e. temperature 0) is used.
All numbers are the result of a single run.

\subsection{Implementation of Explain and Clarify Tools}
\label{sec:tools_implementation}

The implementation of clarification tools in our system enables the verification agent to gather additional information when evaluating a claim. 

\begin{itemize}
    \item \action{explain} prompts the LLM to provide background information about the topic query. As shown in Figure~\ref{lst:generate_background_info}, this action instructs the LLM to synthesize its existing knowledge about the topic in relation to the claim being evaluated.
    \item \action{clarify} disambiguates entities by first retrieving relevant information and then using the language model to explain differences based on that retrieved information. The number of retrieved documents per each clarify is 10. As shown in Figure~\ref{lst:generate_entity_report} and~\ref{lst:generate_entity_report_2}, we prompt the language model to analyze the retrieved results to resolve ambiguities by explain the differences.
\end{itemize}

\subsection{Prompts for Systems}

We list all the prompts used for implementing the agent system below. Figure ~\ref{lst:claim_extraction} is our prompt for extracting atomic facts from Wikipedia articles, whereas Figures ~\ref{lst:generate_background_info}, ~\ref{lst:generate_entity_report}, ~\ref{lst:generate_entity_report_2}, ~\ref{lst:generate_entity_report_3},~\ref{lst:verifier}, ~\ref{lst:verifier_2}, ~\ref{lst:verifier_3}, and ~\ref{lst:controller} are the tools available to the \system agent.
In each prompt, \texttt{\# input} and \texttt{\# output} denote the boundaries of few-shot examples used, if any.

\begin{figure*}[htb]
\begin{lstlisting}
# instruction
You are an expert fact extractor tasked with identifying and listing atomic facts from a given text. Your goal is to produce a comprehensive list of facts that are explicitly stated or directly inferrable from the provided information.

Instructions:
1. Read the title and text carefully.
2. Extract all atomic facts from the information provided. An atomic fact is a single, indivisible piece of information that cannot be broken down further without losing its meaning or accuracy.
3. Include only facts that are explicitly stated or can be directly and unambiguously inferred from the text.
4. Do not add any external knowledge or assumptions not present in the given information.
5. Ensure that each fact is self-contained and can be independently fact-checked.

Before providing your final list of facts, break down your fact extraction process in <fact_extraction_process> tags. This will help ensure a thorough and accurate extraction of facts.

In your fact extraction process, follow these steps:
1. Identify key topics or themes from the title and text.
2. For each topic/theme, list explicit facts from the text.
3. Consider potential inferences that can be directly drawn from the explicit facts, and evaluate their validity.
4. Evaluate each fact (explicit and inferred) for atomicity and self-containment.
5. Categorize facts by topic/theme.
6. Cross-reference each fact with the original text to ensure accuracy.
7. Review the list to ensure no redundant or overlapping facts are included.

After your analysis, provide your final list of facts, with each fact on a new line.

Example output structure:

<fact_extraction_process>
[Your detailed fact extraction process, following the steps outlined above]
</fact_extraction_process>

<facts>
[Fact 1]
[Fact 2]
[Fact 3]
...
</facts>

# input
Here is the title and text you need to analyze:

<title>
{{ full_title }}
</title>

<text>
{{ text }}
</text>
\end{lstlisting}

\caption{Fact Extraction Prompt}
\label{lst:claim_extraction}
\end{figure*}

\begin{figure*}[htb]
\begin{lstlisting}
# instruction
You will be given a topic, and a Wikipedia passage where the topic is mentioned. Your task is to write a self-contained paragraph explaining technical or domain-specific terms in the topic. Your goal is to provide background information on the given topic for people who are unfamiliar with it. If a term, event or concept in the topic has multiple interpretations or meanings, list all plausible ones.

# input
Topic: Infanta Amalia
Wikipedia article: Infanta Amalia of Spain
Infanta Amalia of Spain (Spanish: Amalia de Borbon y Borbon-Dos Sicilias; 12 October 1834 - 27 August 1905) was the youngest daughter of Infante Francisco de Paula of Spain. Her eldest brother, Francisco de Asis, married Queen Isabella II of Spain, who was Amalia's first cousin.

# output
"Infanta Amalia" refers to a title and name in Spanish and Portuguese contexts. "Infanta" is a title used in Spain and Portugal for the daughters of a monarch who are not heir apparent, similar to "princess" in English. "Amalia" is a given name. Therefore, "Infanta Amalia" would refer to a princess named Amalia within a royal family in Spain or Portugal.

# input
Topic: The Great Gatsby
Wikipedia article: The Great Gatsby
It was also performed in the summer of 2012 at the Aspen Music Festival and School. It was performed at Seagle Festival in Schroon Lake, NY in the summer of 2018.

# output
"The Great Gatsby" here likely to a musical adaptation, play, opera, or other performance based on the novel "The Great Gatsby" by F. Scott Fitzgerald. The novel is a classic work of American literature published in 1925. The performances mentioned in the passage are likely adaptations of the novel for the stage or other artistic mediums.

# input
Topic: {{ topic }}
Wikipedia article: {{ claim.context_block.full_title }}
{{ claim.context_block.content }}
\end{lstlisting}

\caption{Generate Background Information Prompt}
\label{lst:generate_background_info}
\end{figure*}

\begin{figure*}[htb]
\begin{lstlisting}
# instruction
You will be given an entity and a Wikipedia paragraph where it is mentioned.
You will also be provided with a list of search results that may contain information about the entity, and other similar entities.
Your task is to write a self-contained paragraph explaining the differences between entities with similar names in the search results.
Entities with similar names might lead to confusion, and the goal here is to disambiguate them. Pay attention to the following:

- People with the same last name, but different first names. Or People with the same name but different professions or time periods.
- Events with the same name but different years or locations. For example, "The Olympics" could refer to the winter or summer games, or games held in different years.
- Organizations with similar names but different purposes or locations.
- etc.

# input
Entity: members of the royal family of Spain named Amalia

[1] Title: Infanta Maria Amalia of Spain
Maria Amalia, Infanta of Spain (9 January 1779 in Madrid - 22 July 1798 in Madrid), was a Spanish princess. She was a daughter of King Charles IV of Spain, in 1795, she married her uncle Infante Antonio Pascual of Spain.

[2] Title: Infanta Amalia of Spain > Childhood
She was born at the royal Palace of Madrid on 12 October 1834 as the eleventh child and sixth daughter of Infante Francisco de Paula of Spain, younger brother of King Fernando VII of Spain, and his wife, Princess Luisa Carlota of Bourbon-Two Sicilies. Infanta Amalia's mother was the niece of her father since her maternal grandmother, Infanta Maria Isabella of Spain, was the elder sister of Infante Francisco de Paula.

[3] Title: Infanta Maria Amalia of Spain > Early life
Born at the Royal Palace of El Pardo, Maria Amalia was the second surviving daughter of King Carlos IV of Spain (1748-1819) and his wife Maria Luisa of Parma (1751-1819), a granddaughter of Louis XV of France.

[4] Title: Infanta Amalia of Spain
Infanta Amalia of Spain (Spanish: Amalia de Borbon y Borbon-Dos Sicilias; 12 October 1834 - 27 August 1905) was the youngest daughter of Infante Francisco de Paula of Spain. Her eldest brother, Francisco de Asis married Queen Isabella II of Spain, who was Amalia's first cousin. She was one of only two of five sisters who made a royal marriage. In 1865 she married Prince Adalbert of Bavaria, a son of King Ludwig I of Bavaria. Upon her marriage she moved to Munich, where she spent the rest of her life. However she remained attached to her native country and was instrumental in arranging the marriage of her eldest son Prince Ludwig Ferdinand of Bavaria with her niece Infanta Paz of Spain.

[5] Title: Infanta Amalia of Spain > Later life and death
Although Infanta Amalia lived for the rest of her life in Munich, she remained attached to her native country. She visited Spain often and her eldest son Prince Ludwig Ferdinand of Bavaria was born at the royal palace of Madrid. She spent the winters at the residence of Munich and the summers at Nymphenburg Palace. Her husband died in 1875; Amalia outlived him by thirty years. Amalia maintained her affiliation with Spain in the next generation. All of her five children spoke Spanish fluently and she encouraged her son Ludwig Ferdinand to marry her niece and goddaughter Infanta Maria de la Paz of Spain. The couple married in 1883.

\end{lstlisting}

\caption{Generate Entity Report Prompt}
\label{lst:generate_entity_report}
\end{figure*}

\begin{figure*}[htb]
\begin{lstlisting}
# output
There are two entities with similar names.
    1. "Infanta Amalia of Spain": Infanta Amalia of Spain (Spanish: Amalia de Borbon y Borbon-Dos Sicilias; 12 October 1834 - 27 August 1905) was the youngest daughter of Infante Francisco de Paula of Spain.
    2. "Infanta Maria Amalia of Spain": Maria Amalia, Infanta of Spain (9 January 1779 in Madrid - 22 July 1798 in Madrid), was a Spanish princess. She was a daughter of King Charles IV of Spain, in 1795, she married her uncle Infante Antonio Pascual of Spain.

These two individuals seem to be separate entities, but may be relatives.

# input
Entity: Antoine Emile Henry Labeyrie

[1] Title: Antoine Emile Henry Labeyrie
Antoine Emile Henry Labeyrie (born 12 May 1943) is a French astronomer, who held the Observational astrophysics chair at the College de France between 1991 and 2014, where he is currently professor emeritus. He is working with the Hypertelescope Lise association, which aims to develop an extremely large astronomical interferometer with spherical geometry that might theoretically show features on Earth-like worlds around other suns, as its president. He is a member of the French Academy of Sciences in the Sciences of the Universe (sciences de l'univers) section. Between 1995 and 1999 he was director of the Haute-Provence Observatory.

[2] Title: Galluis > Notable residents
Antoine-Germain Labarraque (1777 - 1850) was a French chemist and pharmacist, notable for formulating and finding important uses for "Eau de Labarraque" or "Labarraque\'s solution", a solution of sodium hypochlorite widely used as a disinfectant and deodoriser. He died in Gallius on 9 December 1850.

[3] Title: Antoine Lavoisier
Antoine-Laurent de Lavoisier (26 August 1743 - 8 May 1794), also Antoine Lavoisier after the French Revolution, was a French nobleman and chemist who was central to the 18th-century chemical revolution and who had a large influence on both the history of chemistry and the history of biology.

[4] Title: Antoine Germain Labarraque
Antoine Germain Labarraque (28 March 1777 - 9 December 1850) was a French chemist and pharmacist, notable for formulating and finding important uses for "Eau de Labarraque" or "Labarraque\'s solution", a solution of sodium hypochlorite widely used as a disinfectant and deodoriser.

[5] Title: Antoine Germain Labarraque
| Antoine Germain Labarraque | |
| --- | --- |
| Portrait of Labarraque | |
| Born | (1777-03-28) Oloron-Sainte-Marie, Pyrenees-Atlantiques, France |
| Died | 9 December 1850 (1850-12-09) (aged 73)near Paris, France |
| Nationality | French |
| Education | College of Pharmacy, Paris |
| Occupation(s) | chemist and pharmacist |
| Known for | using sodium hypochlorite as a disinfectant and deodoriser |
| Parents | * Francois Labarraque (father) * Christine Sousbielle (mother) |
\end{lstlisting}

\caption{Generate Entity Report Prompt (Continued)} \label{lst:generate_entity_report_2}
\end{figure*}

\begin{figure*}[htb]
\begin{lstlisting}
# output
There are multiple notable French scientists with similar names beginning with "Antoine".
  1. Antoine Emile Henry Labeyrie (born 1943) is a French astronomer and professor emeritus who held the Observational astrophysics chair at the College de France.
  2. Antoine-Laurent de Lavoisier (1743-1794) was a French nobleman and chemist central to the 18th
  3. Antoine Germain Labarraque (1777-1850) was a French chemist and pharmacist known for developing "Labarraque's solution," a sodium hypochlorite disinfectant.
While these individuals share similar first names and French nationality, they worked in different fields and time periods.

# input
Entity: {{ entity_name }}
Original article: {{ claim.context_block.full_title }}
{{ claim.context_block.content }}

{{ search_results.to_string() }}
\end{lstlisting}

\caption{Generate Entity Report Prompt (Continued)} \label{lst:generate_entity_report_3}
\end{figure*}

\begin{figure*}[htb]
\begin{lstlisting}
# instruction
Determine if a claim extracted from a Wikipedia paragraph is inconsistent with any of the provided documents. A claim is deemed inconsistent when at least one document contains information that directly contradicts it. If no such contradiction exists - even when the documents do not explicitly support the claim - the claim is considered consistent.

Step-by-Step Instructions:

1. Identify the Claim
Definition: A brief statement directly extracted from a Wikipedia paragraph.
Note: The full meaning of the claim might require context provided by the original paragraph.

2. Review the Documents
Definition: Passages, tables, or pieces of text retrieved from Wikipedia.
Task: Ignore documents that are clearly irrelevant to the claim.
Focus on finding any document that might contain information in clear conflict with the claim.

3. Consider Clarifications

Definition: Additional background information provided to clarify ambiguous terms or entities.
Task: Use clarifications to distinguish between similar or similarly named entities.
Important: Do not use clarifications to support or contradict the claim directly - they serve only to clear up ambiguities.

4. Assess for Inconsistencies

Definition of Inconsistency:
The claim is inconsistent if at least one document provides information that contradicts it.
Conversely, if no document provides conflicting information, the claim is considered consistent.
Measurement: Assign an inconsistency score between 0 (fully consistent) and 1 (completely inconsistent).
Intermediate scores indicate varying degrees of uncertainty or partial conflict.

5. Common Scenarios & Examples

Example 1: Clear Inconsistency

Claim: "The capital of Thailand is Bangkok."
Document: "The capital of Thailand is Phuket."
Reasoning: A country typically has one capital. The document contradicts the claim by listing a different city, yielding a high inconsistency score (e.g., 0.8-0.9).

Example 2: Apparent Inconsistency Resolved by Entity Equivalence (Minor Inconsistency)

Claim: "The capital of Thailand is Bangkok."
Document: "The capital of Thailand is Krung Thep Maha Nakhon."
Additional Background: It is widely accepted that Bangkok and Krung Thep Maha Nakhon refer to the same city.
Reasoning: Although the names differ, they reference the same location; thus, the claim is largely consistent (e.g., inconsistency score around 0.2-0.4).
Note: If an explicit clarification were provided stating the equivalence, the score would be 0.

Example 3: Misplaced Terms Causing Inconsistency

Claim: "The capital of Thailand is Bangkok."
Document: "The capital of Bangkok is Thailand."
\end{lstlisting}

\caption{Verifier Prompt} \label{lst:verifier}
\end{figure*}

\begin{figure*}[htb]
\begin{lstlisting}
Reasoning: The document seems to mix up entities by stating that Bangkok is a country. With no supporting evidence that this is a mere typo or misinterpretation, the conflict earns a high inconsistency score (e.g., around 0.9).

Example 4: Inconsistent Translational Variants

Claim: "The 'Song is Universal' won the Best Modern Rock Song award at the 2010 Korean Music Awards."
Document: "The Best Modern Rock Song award at the 2010 Korean Music Awards was given for 'Universal Song.'"
Additional Clarification: "Bangkok only refers to a city in Thailand, not elsewhere." (Not directly applicable here but shows how clarifications work.)
Reasoning: Although the song likely is the same, the differing English translations ("Song is Universal" vs. "Universal Song") introduce an inconsistency, resulting in a moderately high inconsistency score (e.g., around 0.8).

Example 5: No Conflict (Consistency)

Claim: "Stress is harmful to health, as mentioned in the medical literature."
Document: "Stress is necessary for growth and development, pushing limits, enhancing learning, and building resilience."
Reasoning: The document discusses the beneficial aspects of acute or eustress compared to chronic stress, which is what the claim addresses. Since these are two different perspectives on stress, there is no contradiction - the claim is consistent (inconsistency score 0).

Final Decision

After review, provide the inconsistency score for the claim:
0: Fully consistent; no document contradicts the claim.
Between 0 and 1: Partial or potential inconsistencies.
1: Fully inconsistent; at least one document directly contradicts the claim.

# input
<claim>
Title: {{ claim.context_block.full_title }}
{{ claim.context_block.content }}

You should only focus on the aspect of this paragraph related to: "{{ claim.claim_text }}"
</claim>

Read the following clarifications about the claim:
<clarifications>
{%
[{{ loop.index }}] {{ clarification }}

{%
</clarifications>

Now, read through the documents below and look for any information that conflicts with the claim:
<documents>
{%
[{{ loop.index }}] {{ document }}

{%
</documents>
\end{lstlisting}

\caption{Verifier Prompt (Continued)} \label{lst:verifier_2}
\end{figure*}

\begin{figure*}[htb]
\begin{lstlisting}
Now, you need to analyze the documents and clarifications to determine an inconsistency score that represents your confidence that the claim is inconsistent with the documents.

First, rephrase the claim to be more specific by:
1. Incorporating context from the Wikipedia article title and content
2. Preserving the original meaning, but making corrections if the claim appears to be misrepresented or incorrectly paraphrased from the claim's context.

<claim_with_context>
[Provide the claim that incorporates the context from the Wikipedia article title and content here.]
</claim_with_context>

Based on the claim with context, present your full analysis and arguments:

<analysis>
[Provide a detailed analysis by:
1. Carefully examining the claim and documents for any contradictions or inconsistencies, look through examples above if there is concept similar to your case
2. Highlighting specific documents where information directly conflicts with the claim
3. Making sure that these documents are relevant to the claim. Some documents may contain the same entities as the claim, but they are not relevant to the claim, given the context
4. Exploring multiple interpretations of the claim's meaning and implications
5. Considering edge cases and ambiguities that could affect the analysis
6. Referencing relevant examples from above (translations, time-related issues, ordering) to strengthen your reasoning
7. Explaining your confidence level in identifying any inconsistencies found]

</analysis>

Based on your analysis, provide an inconsistency score from 0 to 1, where:
- 0 indicates the claim is completely consistent with all of the documents
- 1 indicates the claim is completely inconsistent with at least one of the documents
- Values between 0 and 1 represent varying degrees of uncertainty

<inconsistency_score>
[A single float from 0 to 1]
</inconsistency_score>
\end{lstlisting}

\caption{Verifier Prompt (Continued)} \label{lst:verifier_3}
\end{figure*}

\begin{figure*}[htb]
\begin{lstlisting}
# instruction
You will be given a "claim" statement extracted from a Wikipedia paragraph.
Your task is to conduct a thorough investigation on the entire Wikipedia (except the article where the claim comes from) to find any factual inconsistencies with this claim.
As you conduct your investigation, you may come across articles that support the claim. However, you should continue searching for inconsistencies that might exist in other places. Inconsistencies might appear in subtle or indirect ways.

You will conduct your investigation in multiple steps. At each step, you should think about the information you have gathered so far, and choose one of the following actions based on it:

- `explain(topic: str) -> str`: Use this action to understand the basics of a specific term or concept you encounter, for example a technical term or the rules of a sport.

- `clarify_entity(entity_name_and_description: str) -> str`: Use this action to get a report on an entity (person, organization, event etc.) to clarify other entities with similar names. This will help you properly differentiate similar-sounding entities when researching inconsistencies. For example, clarify_entity("WW III wrestling event") will explain all potential events with similar names, or the same event in different years.

- `search_wikipedia_outside_claim_article(question: str) -> list`: Use this action to explore Wikipedia.

- `report_inconsistency(evidence: str)`: If at any point you are certain that you have found an inconsistency, use this action to report it. Evidence should be a short sentence that describes the inconsistency. Once you report an inconsistency, a human will review it and provide feedback.

# input
Here is the claim to find inconsistencies with:
{{ claim.claim_text }}

Here is more context about the claim for your reference:
Title: {{ claim.context_block.full_title }}
{{ claim.context_block.content }}

{%
Actions you have taken so far:

{{ action_history}}
{%
\end{lstlisting}

\caption{Controller Prompt}
\label{lst:controller}
\end{figure*}